\documentclass{article}

    \PassOptionsToPackage{numbers, compress}{natbib}

  \usepackage[main, final]{neurips_2026}

\usepackage[utf8]{inputenc} %
\usepackage[T1]{fontenc}    %
\usepackage{hyperref}       %
\usepackage{url}            %
\usepackage{booktabs}       %
\usepackage{amsfonts,amsmath,amssymb} %
\usepackage{nicefrac}       %
\usepackage{microtype}      %
\usepackage{xcolor}         %
\usepackage{graphicx}
\usepackage{subcaption}
\usepackage{wrapfig}
\usepackage{xspace}
\newcommand{\bench}{\textsc{ACED}\xspace}                 %
\newcommand{\query}{\textsc{ACED-Query}\xspace}           %
\newcommand{\repair}{\textsc{ACED-Repair}\xspace}
\newcommand{\acedqual}{\textsc{Qual}\xspace}              %
\newcommand{\acedcore}{\textsc{Core}\xspace}              %
\newcommand{\fixrate}{fix rate\xspace}
\newcommand{\B}{\mathrm{B}}                               %
\newcommand{\U}{U}
\newcommand{\doop}{\mathrm{do}}

\usepackage{array}
\newcolumntype{P}[1]{>{\raggedright\arraybackslash}p{#1}}
\usepackage{multirow}
\usepackage[capitalise,noabbrev]{cleveref}
\graphicspath{{figures/}}

\title{\bench: Evaluating How LLM Agents Acquire and Act on Causal Evidence}

\author{%
  Yiwei Chen\thanks{Email: \texttt{vic010@ucsd.edu}} \\
  UC San Diego
  \And
  Amila Weerasinghe \\
  Amazon
  \And
  Siddharth Biswal \\
  Amazon
  \And
  Utkrisht Rajkumar \\
  Amazon
  \AND
  Olcay Boz \\
  Amazon
  \And
  Taylor Berg-Kirkpatrick \\
  UC San Diego
  \And
  Rose Yu \\
  UC San Diego
}

\begin{document}

\maketitle

\begin{abstract}
Large language models are increasingly used as scientific assistants, yet we still lack reliable ways to evaluate whether they can investigate a system rather than merely answer a prompt.
A useful evaluation must let agents choose experiments, interpret finite evidence, and make causal decisions, while preserving objective ground truth.
We introduce \bench{} (\textbf{A}ctive \textbf{C}ausal \textbf{E}vidence and \textbf{D}ecision Benchmark), a neuro-symbolic benchmark that separates the readable scientific surface from the formal causal world: natural-language scenarios describe the system, while a hidden causal model defines the data, the effect of every intervention, and the oracle answers.
In \repair, agents diagnose a failing production system: within 15 experiments they must decide which controls to test, read confounded and filtered data, and recommend a fix, which we re-simulate on a hidden true objective.
In \query, agents answer causal questions and make constrained causal decisions in worlds backed by categorical probabilistic graphical models.
Acceptance on \repair ranges from 43\% to 0\% across 12 LLMs, while strong agents approach ceiling on \query's qualification questions.
Tracing every experiment shows where investigations stop: most open-weight models test a true lever on its own in fewer than half of the worlds, the strongest models test the right levers but often miss the working dose or combination, and agents that do run a working fix almost always recommend it.
No model reliably repairs systems with two independent causes or systems that need a subgroup-specific policy.
These results suggest that reliable AI co-scientists need explicit support for choosing experiments, keeping track of evidence, and composing interventions.
Code and data are available at \url{https://github.com/Rose-STL-Lab/ACED-Bench}.
\end{abstract}

\section{Introduction}
\label{sec:intro}

Large language models (LLMs) are increasingly used as scientific assistants: they read papers, write code, analyze data, and propose hypotheses.
If such systems are to become reliable AI co-scientists, we need to evaluate how they investigate as well as what they answer.
Scientific investigation requires deciding what would be informative to observe, choosing experiments, analyzing data, revising beliefs, and committing to a conclusion when the evidence is sufficient.
This is especially important for causal questions, where patterns in passively observed data can differ from the effects of actively changing a system.
A strong scientific agent must therefore know how to reason about causality and which experiment would make a causal claim testable.

Existing benchmarks capture parts of this problem, but rarely combine them.
Causal-reasoning benchmarks test concepts such as intervention, confounding, and counterfactuals; statistical-reasoning benchmarks test inference from data; and scientific-agent benchmarks test tool use and multi-step investigation.
What remains difficult is evaluating whether an agent can design useful experiments and make causal decisions in a world that is both readable in natural language and formally consistent.
Such a benchmark must support repeated observation and intervention, provide coherent outcomes across experiments, and retain objective ground truth.

We introduce \bench{}, a neuro-symbolic benchmark for evaluating whether LLM agents can investigate causal systems through experimental evidence.
Each task is set in a natural-language scientific world backed by a hidden causal model: agents may request observational data, run interventions, and write code to analyze the resulting finite samples under a fixed budget.
Because the full model is known to the evaluator, gold answers are computed programmatically rather than manually labeled, for every experiment an agent might run and every answer it might give.

\begin{figure}[t]
    \centering
    \includegraphics[width=\linewidth]{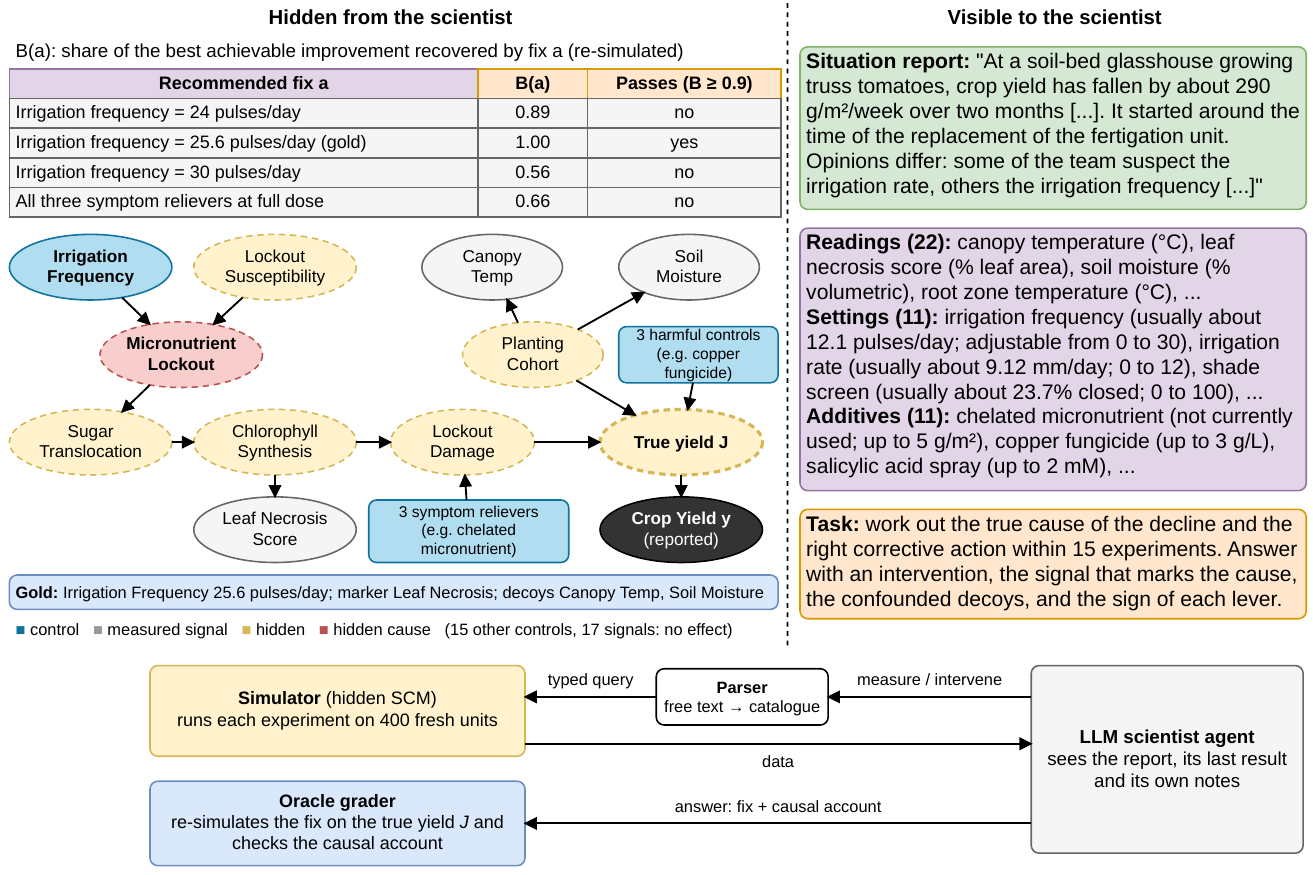}
    \caption{
    One \repair world (agronomy, instrument-only structure) and the interaction loop.
    The hidden SCM (top left) defines the outcome of every experiment and the benefit $B(a)$ of any fix; the agent sees only the situation report, the catalogue of readings, settings and additives, and the task (top right).
    The agent requests experiments in free text, a fixed parser maps them to the catalogue, and the simulator returns data; the oracle grades the final answer on the hidden model.
    \query uses the same design with categorical PGMs (\cref{fig:query_overview}).
    }
    \label{fig:overview}
\end{figure}

Our main suite, \repair, poses the problem a process scientist faces (\cref{fig:overview}).
A production system's output has dropped, and the agent has 15 experiments to find the cause and recommend a fix.
Each world hides one of nine causal structures, such as confounding, selection, a surrogate outcome, or a treatment that helps only one subgroup, so a fix chosen from correlations without testing usually fails.
The agent must decide which of about 25 controls to test, in which direction and at which dose, and must read data that are confounded or filtered.
We grade the recommended fix by re-simulating it on a hidden true objective, and the agent's causal account by a separate check.
A second suite, \query, uses hidden probabilistic graphical models with categorical variables: agents answer causal questions and make constrained causal decisions, such as choosing a treatment that helps without harming a protected group.
In \query the question names the causal contrast or decision, while in \repair the agent must work out what to test; together the two suites separate using evidence from deciding which evidence to collect.

We evaluate 12 LLMs on \repair and 17 on \query.
\repair separates frontier models widely, while strong agents approach ceiling on \query's qualification questions (\cref{sec:results}).
Because every experiment runs on the hidden model, we can trace how far each investigation gets (\cref{sec:analysis}).
Models stop at different points, from choosing which control to test to finding a dose or combination that works.
Protocol ablations show that keeping track of evidence across turns matters for some models, and repairs that need fixes for two independent causes or a subgroup-specific policy remain hard for every model.

Our contributions are:
\begin{itemize}
    \item \bench{}, a design for evaluating LLM agents on active causal investigation in which natural-language scenarios are backed by hidden causal models that define every observation, intervention and oracle answer, with two suites: \repair (diagnosis and repair under nine causal structures) and \query (causal questions and decisions).
    \item An evaluation of 12 LLMs on \repair and 17 on \query, with repeated runs for the leading models and classical procedures as reference points.
    \item An analysis of how agents choose experiments: a stage-by-stage trace of every \repair investigation, an evidence ledger for \query, and protocol ablations on memory and budget.
\end{itemize}

\section{Related work}
\label{sec:related}

\paragraph{Agentic scientific and data-analysis benchmarks.}
Recent benchmarks increasingly evaluate LLMs as agents that interact with tools, environments, or data sources.
ScienceWorld and DiscoveryWorld test scientific exploration in interactive text environments, with DiscoveryWorld requiring hypothesis formation, experimentation, and explanation~\citep{wangScienceWorldYourAgent2022a,jansenDiscoveryWorldVirtualEnvironment2024}.
DiscoveryBench, BLADE, DA-Bench, ScienceAgentBench, and KramaBench evaluate data-driven investigation, code execution, and data-to-insight workflows~\citep{majumderDiscoveryBenchDataDrivenDiscovery2024,guBLADEBenchmarkingLanguage2024,hu2024infiagentdabench,chenScienceAgentBench2025,laiKramaBenchBenchmarkAI2026}, and end-to-end systems and environments evaluate automated research~\citep{luAIScientistFully2024b,yamadaAIScientistv2WorkshopLevel2025b,garikaparthiResearchGym2026}.
These benchmarks establish that scientific-agent evaluation should include planning, tool use, and iterative interaction; \bench{} adds exact observational and interventional semantics to that setting.

\paragraph{Causal and statistical reasoning benchmarks.}
Causal-reasoning benchmarks test whether LLMs understand concepts such as association, intervention, confounding, and counterfactuals~\citep{jinCLadderAssessingCausal2023,wangCausalBenchComprehensiveBenchmark2024,jinCanLargeLanguage2024,duIceCreamDoesnt2026}, and InterveneBench evaluates intervention reasoning and study design for real social systems~\citep{shiInterveneBenchBenchmarkingLLMs2026}.
Statistical-reasoning benchmarks evaluate probability, uncertainty, calibration, and data analysis~\citep{liuAreLLMsCapable2024,paruchuriWhatAreOdds2024,nafarReasoningUncertainText2025,zhuAreLargeLanguage2024,rendaOpenEstimateEvaluatingLLMs2026,luStatEvalComprehensiveBenchmark2025}.
CauSciBench is especially close in its focus on causal-scientific reasoning, but the relevant evidence in these benchmarks is usually given directly as text, a graph, or a fixed dataset~\citep{acharyaCauSciBenchAssessingLLM2025}.

\paragraph{LLMs for causal discovery and experimental design.}
Several works use LLMs to assist causal discovery, for example by supplying background knowledge, orienting edges, reducing equivalence classes, or guiding intervention selection~\citep{longCausalDiscoveryLanguage2023,takayamaIntegratingLargeLanguage2025,abdulaalCausalModellingAgents2024,liCanLargeLanguage2025}.
Interactive benchmarks evaluate LLM agents as experimenters directly: BoxingGym evaluates experimental design and model discovery against hidden probabilistic models~\citep{gandhiBoxingGym2025}, PhysGym and NewtonBench ask agents to probe simulated environments to recover physical laws~\citep{chenPhysGym2025,zhengNewtonBench2025}, and Auto-Bench evaluates iterative discovery of hidden causal structure with an oracle~\citep{chenAutoBenchAutomatedBenchmark2025}.
On gene-perturbation and molecular-property design tasks, \citet{guptaLLMs2025} find LLM agents insensitive to permuted feedback and behind classical bandit and Bayesian optimisation methods.

\paragraph{Diagnosis and root-cause analysis.}
OpenRCA evaluates root-cause localisation from software telemetry~\citep{xuOpenRCA2025}, \citet{kimWhy2026} analyse why agents fail at this task, and sequential diagnosis benchmarks let agents choose questions and tests before committing to a clinical diagnosis~\citep{noriSequential2025}.
In \repair, agents also intervene on the system, and the score is whether the recommended repair works.

\paragraph{Active causal learning and intervention design.}
Our setting is grounded in causal inference and PGMs~\citep{pearl2009causality,koller2009probabilistic}.
It is also related to active causal learning, where interventions are selected to identify structure or reduce uncertainty~\citep{he2008active,hauserCharacterizationGreedyLearning2012,tothActiveBayesianCausal2022,zhangActiveLearningOptimal2023}.
These methods assume an explicit causal model or posterior and optimise an acquisition objective over interventions~\citep{zhangActiveLearningOptimal2023}, while \bench{} evaluates general-purpose LLM agents that interact through natural language, sampling APIs, and optional code execution.

\paragraph{Trajectory-level evaluation and evidence management.}
Final-answer accuracy can hide important differences between agents that make the same mistake for different reasons.
Process supervision rewards correct intermediate reasoning steps~\citep{lightmanLetsVerifyStep2023}, and agent designs such as ReAct and Reflexion interleave reasoning with actions and self-reflection~\citep{yaoReActSynergizingReasoning2023b,shinnReflexionLanguageAgents2023}.
Trajectory-level evaluations locate where agent runs go wrong~\citep{kimFinalAnswerEvaluating2025,fanAgentProcessBenchDiagnosingStepLevel2026,barkeAgentRx2026}, and long-horizon research environments identify hard limits from context length among agent failure modes~\citep{garikaparthiResearchGym2026}.

\paragraph{Concurrent work.}
Concurrent with this work, several benchmarks have also placed agents in hidden causal or scientific environments, including CausaLab~\citep{yangCausaLab2026}, ACDB~\citep{debActive2026}, CausalGame~\citep{chenCausalGame2026}, ActiveSciBench~\citep{kabraLLMAutoSciLab2026}, Autonomous Science Bench~\citep{wenAutonomous2026}, and CausalDS~\citep{lebanCausalDS2026}.

\section{\bench}
\label{sec:design}

\subsection{Hidden-world interaction}
\label{sec:design:shared}

\bench{} evaluates LLM agents as sample-based causal investigators in hidden causal worlds.
Each world is a tuple
\[
    W = (\mathcal{M}, \mathcal{V}, s, \mathcal{Q}, \mathcal{N}),
\]
where $\mathcal{M}$ is a causal model over a set of variables $V$ (a Bayesian network over categorical variables in \query, a structural causal model over continuous variables in \repair), $\mathcal{V}$ contains variable names, descriptions, units, and value sets, $s$ is a natural-language scientific scenario, $\mathcal{Q}$ is the task, and $\mathcal{N}\subseteq V$ is the set of non-intervenable variables.
The evaluator has access to the full model.
The agent observes only the scenario, variable metadata, the task, and its interaction history.
At each turn, the agent may request observational samples from $P_W(V)$, request interventional samples from $P_W(V\mid \doop(X=x))$, run code when permitted, or submit a final answer.

\bench{} makes the scientific story readable without letting language determine the answer.
The language layer supplies variable names, value descriptions, scenarios, and task text; a programmatic backend constructs the causal model, samples its parameters, defines interventions, validates the world, and computes the oracle answer.
The agent reads a domain-themed scenario, while the evaluator retains exact causal ground truth, including the outcome of interventions the agent never ran.
This separation also makes the benchmark scalable: new domains, graph sizes, and structure instances can be generated and filtered without manual answer labeling.

A fixed parser maps the agent's free-text requests to typed simulator queries, or rejects them.
The parser performs no causal reasoning, and deterministic checks enforce query validity and intervention constraints, so every agent's requests are interpreted by the same rules; \cref{app:parser} reports the parser's reliability in both suites.

\subsection{\repair: diagnosis and repair}
\label{sec:design:repair}

An \repair world is an SCM $\mathcal{M}$ of a production system, such as a greenhouse, a bioreactor, a data centre, or a clinical ward, in one of ten domains.
A hidden cause has reduced a true objective $J$, which the agent never observes; it observes a reported outcome $y$ that tracks $J$ except where the structure says otherwise.
Each world has about 25 controls and 26 measurable signals.
A control is either a \emph{lever}, which acts on the cause, a symptom reliever, which raises the outcome without addressing the cause, a harmful control, or an inert one.
The situation report lists every signal and control with its units, usual level and range, and describes when the loss began.

Names, units, ranges, and the story are realistic for the domain, while the causal role of each control is assigned at random.
Domain knowledge is deliberately unreliable as a guide to causal structure, so the agent has to establish the cause by experiment.

Each world instantiates one of nine causal structures (\cref{tab:structures}).
The test set has 540 worlds, 60 per structure.
Every model is evaluated on a 54-world evaluation set drawn from it, six worlds per structure.

\begin{table}[t]
\centering
\small
\caption{
The nine causal structures of \repair.
Each defeats a different fix chosen from correlations without testing.
The last three columns give the \fixrate{} (\%): Claude Opus 5.5 on all 540 test worlds, and GPT-6 Astra and GPT-6.1 Sol on the evaluation set (means over three runs).
}
\label{tab:structures}
\begin{tabular}{@{}P{2.0cm}P{3.9cm}P{2.8cm}rrr@{}}
\toprule
Structure & What misleads a correlational fix & What the correct fix needs & Opus & Astra & Sol \\
\midrule
Confounded chain & a confounded correlate of the outcome looks causal & act on the upstream setting & 53 & 61 & 61 \\
Collider / selection & the observational record is filtered by a selection mechanism & compare interventions, not the filtered record & 58 & 56 & 50 \\
Confounded reversal & the observational sign is opposite to the needed direction & move the lever the counter-intuitive way & 38 & 56 & 39 \\
Instrument only & the cause is hidden and reachable only through one setting & find and dose that setting & 65 & 83 & 61 \\
Surrogate trap & one control moves the reported outcome but not the true objective & ignore the reading-only control & 72 & 78 & 72 \\
Dose window & the effect is non-monotone with an interior optimum & bracket the dose & 37 & 72 & 61 \\
Competing causes & two independent causes each explain part of the loss & fix both & 2 & 6 & 0 \\
Synergy pair & two controls act mainly together & combine them & 3 & 61 & 50 \\
Hidden subtype & a treatment helps one subgroup and harms the other & a policy conditioned on a measured marker, plus the fix for a second cause & 0 & 6 & 0 \\
\bottomrule
\end{tabular}
\end{table}

In each turn the agent writes one action.
\emph{Measure} records signals and control settings on 400 units as they currently run, and \emph{intervene} sets any combination of controls to any values on a fresh batch of 400 units.
\emph{Code} runs Python on the data collected so far.
\emph{Answer} submits a recommended intervention, optionally a policy conditioned on a measured marker, and a causal account: the signal that marks the hidden cause, the confounded decoys among the correlates of the outcome, and the effect sign of each lever.
The agent has 15 experiments and 32 turns.

Each turn shows the situation report, the counters, the result of the last action, and the agent's own notes from the previous turn.
Earlier results reach later turns only through these notes or through code over the stored data, so keeping track of evidence is part of the task (\cref{sec:analysis:memory}).

\subsection{\repair scoring}
\label{sec:design:grading}

We score the recommended fix by how much of the oracle fix's improvement it recovers.
Let $\U(a) = \mathbb{E}_{\mathcal{M}}[J \mid \doop(a)]$ be the expected true objective under intervention $a$, and let $a^\star$ be the best intervention found by the oracle's search over single controls and combinations of up to three, with dose refinement (\cref{app:repair}).
The benefit recovered by a recommendation $\hat a$ is
\begin{equation}
    \B(\hat a) = \max\!\left(0,\; \frac{\U(\hat a) - \U(\varnothing)}{\U(a^\star) - \U(\varnothing)}\right),
    \label{eq:benefit}
\end{equation}
so doing nothing scores 0, a harmful recommendation also scores 0, and the oracle's fix scores 1.
The fix passes if $\B(\hat a) \ge 0.9$, which requires the right controls at roughly the right dose; we call the share of passing worlds the \emph{\fixrate}.

The causal account $\hat m$ is scored by a mechanism check $\beta(\hat m)$, the mean of one item for the marker, one for the decoy set (Jaccard overlap with the true decoys), and one per lever for its sign; it passes if $\beta(\hat m) \ge 0.8$ (\cref{app:repair}).
A world is \emph{accepted} if both pass:
\begin{equation}
    \mathrm{accepted} = \mathbf{1}\!\left[\B(\hat a) \ge 0.9\right]\cdot\mathbf{1}\!\left[\beta(\hat m) \ge 0.8\right].
    \label{eq:accepted}
\end{equation}

\subsection{\query: causal questions and decisions}
\label{sec:design:query}

\query worlds are hidden PGMs over categorical variables with 10 to 30 nodes, spanning eight topical domains.
An LLM drafts the variable names, value descriptions, scenarios, and task text; for causal-decision tasks, a decision archetype fixes the semantic roles, graph constraints, validation checks, and the oracle answer.
Worlds with broken graph constraints, small effect margins, or ambiguous oracle answers are rejected; \cref{app:query} shows the generation pipeline and the structural task interface.

The main leaderboard \acedcore{} combines causal ancestry/descendancy list queries, which ask for all causes or all effects of a target variable, with causal-decision questions built around safety, robustness, mediation, satisficing, and invalid-premise archetypes.
\acedqual{} is a separate qualification check with yes/no causal-effect, marginal, and conditional questions; because the question specifies the target contrast, it asks whether an agent can obtain and interpret the relevant evidence once the scientific question has already been posed.
Agents answer within four turns, which makes choosing which experiment to run the binding constraint.
Every active agent answers the same 70 \acedcore{} tasks (10 list queries and 60 decisions) and 50 \acedqual{} questions; the guess-only baseline, which uses no data, is scored on a separate, larger set of 78 \acedcore{} tasks and 150 \acedqual{} questions.

\acedqual{} questions are scored by accuracy, and \acedcore{} by the share of exactly matched lists and correct decisions:
\begin{equation}
    \mathrm{acc}_{\acedcore} = \frac{\#\{\text{exactly matched sets}\} + \#\{\text{correct decisions}\}}{\#\{\text{tasks}\}}.
    \label{eq:core}
\end{equation}
A decision is correct if its interventional effect, computed on the network, meets the archetype's criterion; malformed or unparseable answers and give-ups score 0 (\cref{app:query}).

\section{Results}
\label{sec:results}

\subsection{Evaluation protocol}
\label{sec:protocol}

\paragraph{Models.}
We evaluate 17 models: Claude Opus 5.5, Sonnet 5.5, Haiku 5.5, Opus 4.6 and Sonnet 4.6, GPT-6 Astra and GPT-6.1 Sol, Gemini 3.1 Pro, DeepSeek-V4-Pro, Qwen3.8-27B, Qwen3.6-27B, Qwen3.5-9B, Gemma 4 31B, GPT-OSS-120B, Llama 4 Maverick, and, as earlier-generation references, GPT-4o and Llama 3.3 70B.
\repair uses twelve of them, all except Claude Opus 4.6 and Sonnet 4.6, Qwen3.6-27B, GPT-4o and Llama 3.3 70B.
Every model runs with a long output limit; \cref{app:models} lists the sampling settings.

\paragraph{Agents.}
\query compares four agent designs.
The \textbf{guess-only} agent (Guess) answers immediately without data access.
The \textbf{active reasoning} agent (Active) requests observational or interventional samples and reasons from simulator summaries, but cannot execute code.
The \textbf{single-prompt active-code} agent (Active+Code) can additionally write Python code to analyze sampled CSV files.
The \textbf{modular active-code} agent (Modular) separates planning, code analysis, interpretation, and experiment design into explicit stages.
\repair uses one agent loop for all models (\cref{sec:design:repair}), whose prompt describes the actions, the budget, the memory rule, and the grading criteria.
Prompts are identical across models, except that for Claude Opus 5.5 and Sonnet 5.5 on \query the instruction to write reasoning before each action was removed and the modular agent's first instructions were sent in the user turn, because the provider's content filter rejected the original prompt; Qwen3.5-9B runs \query with thinking off.

\paragraph{Runs and statistics.}
Each model answers each \query task and each world of the evaluation set once; the three leading \repair models run three times, and we report their mean.
Claude Opus 5.5 also runs once on all 540 test worlds.
Intervals are 95\% bootstrap intervals over worlds, and protocol ablations compare paired runs on the same worlds with an exact McNemar test.

\subsection{\repair}
\label{sec:results:repair}

\begin{figure}[t]
    \centering
    \includegraphics[width=0.92\linewidth]{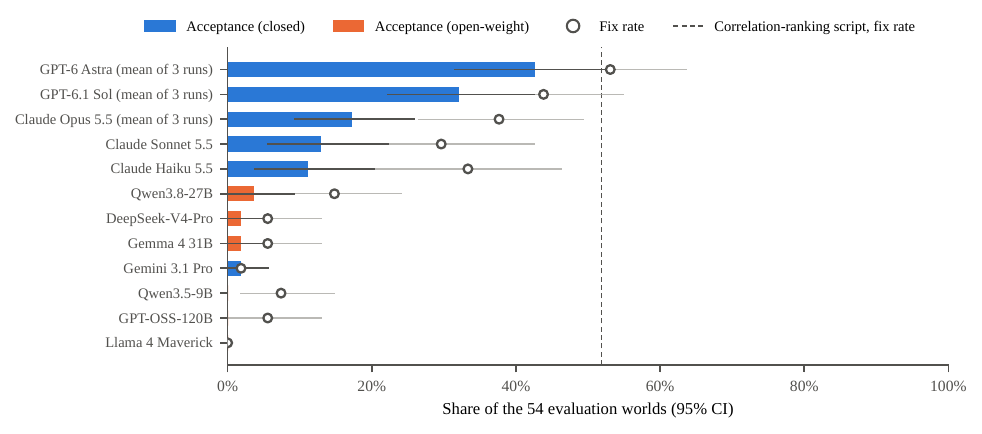}
    \caption{
    \repair on the evaluation set.
    Bars: acceptance (fix and mechanism both pass); circles: \fixrate{} (the recommended fix restores the true objective).
    Dashed line: \fixrate{} of a scripted procedure that ranks controls by their correlation with the outcome and tests them (\cref{sec:analysis:scripts}).
    95\% bootstrap intervals over worlds.
    }
    \label{fig:leaderboard}
\end{figure}

\repair spreads frontier models much further apart than \query and orders them differently (\cref{fig:leaderboard}).
GPT-6 Astra is accepted in 43\% of evaluation worlds, GPT-6.1 Sol in 32\% and Claude Opus 5.5 in 17\%, while every open-weight model and Gemini 3.1 Pro stay below 4\%.
Gemini 3.1 Pro is in the top group of \query and almost never repairs an \repair world: answering a specified causal question well does not guarantee success at investigating a failing system.
On all 540 test worlds, Opus 5.5 fixes 36.5\% of worlds, close to its rate on the evaluation set.

Finding a working fix is easier than explaining it.
For every model that fixes more than 10\% of worlds, acceptance is 10 to 22 points below the \fixrate{}, because agents recommend a working fix with the wrong marker, an incomplete decoy set, or a wrong sign for a lever.

Agents solve many single-cause worlds, and the GPT-6 models also solve many synergy worlds, but no model repairs competing causes or hidden subtypes in more than 6\% of worlds (\cref{tab:structures}).
In competing-cause worlds agents rarely recommend fixes for both causes, and in hidden-subtype worlds Claude Opus 5.5 wrote the exact subgroup policy in three of its test-set episodes but each time omitted the fix for the other cause.

Single runs vary: GPT-6.1 Sol, for example, is accepted in 43\%, 22\% and 31\% of the same worlds in three runs, while the leading group itself is stable across runs (\cref{app:ablations}).

\subsection{\query}
\label{sec:results:query}

\begin{table}[t]
\centering
\small
\caption{\query accuracy by model and agent. Rows are sorted by mean \acedcore{} accuracy over the three active agents, and bold marks the best active-agent cell in each column; Guess is scored on a separate, larger task set (\cref{sec:design:query}); 95\% world-clustered intervals are in \cref{app:query}.}
\label{tab:query}
\setlength{\tabcolsep}{4pt}
\begin{tabular}{@{}lcccc|c@{}}
\toprule
 & \multicolumn{4}{c|}{\acedcore} & \acedqual \\
Model & Guess & Active & Active+Code & Modular & Modular \\
\midrule
Claude Sonnet 5.5 & 0.40 & \textbf{0.86} & \textbf{0.89} & 0.91 & 1.00 \\
Claude Opus 5.5 & 0.49 & 0.83 & 0.86 & \textbf{0.94} & 0.98 \\
Gemini 3.1 Pro & 0.58 & 0.83 & 0.79 & 0.90 & 0.96 \\
Claude Haiku 5.5 & 0.47 & 0.76 & 0.73 & 0.86 & 1.00 \\
Gemma 4 31B & 0.45 & 0.73 & 0.77 & 0.79 & 0.94 \\
Qwen3.8-27B & 0.41 & 0.77 & 0.74 & 0.76 & 0.92 \\
Claude Opus 4.6 & 0.41 & 0.76 & 0.77 & 0.71 & 0.64 \\
DeepSeek-V4-Pro & 0.42 & 0.71 & 0.81 & 0.64 & 0.82 \\
Qwen3.6-27B & 0.44 & 0.71 & 0.70 & 0.76 & 0.98 \\
Claude Sonnet 4.6 & 0.53 & 0.80 & 0.61 & 0.69 & 0.90 \\
Llama 4 Maverick & 0.41 & 0.66 & 0.61 & 0.67 & 0.88 \\
GPT-6.1 Sol & 0.42 & 0.63 & 0.77 & 0.47 & 0.66 \\
GPT-6 Astra & 0.45 & 0.64 & 0.77 & 0.41 & 0.62 \\
Qwen3.5-9B & 0.46 & 0.63 & 0.60 & 0.54 & 0.74 \\
Llama 3.3 70B & 0.38 & 0.64 & 0.37 & 0.60 & 0.88 \\
GPT-OSS-120B & 0.45 & 0.53 & 0.19 & 0.51 & 0.70 \\
GPT-4o & 0.40 & 0.44 & 0.33 & 0.39 & 0.74 \\
\bottomrule
\end{tabular}
\end{table}

Strong agents can use the sampling interface when the relevant contrast is specified.
On \acedqual{}, nine of the 17 models reach at least 0.96 with some agent (\cref{tab:query_full}).

\acedcore{} preserves substantial spread across models and agent designs.
The best cells come from Claude Opus 5.5, Claude Sonnet 5.5 and Gemini 3.1 Pro with the modular agent (0.90--0.94), and bootstrap resampling places each of them in the top three at least 78\% of the time.
Code access helps some models and hurts others, which lose accuracy when the code loop consumes scarce turns.

Two kinds of task remain hard for every agent design.
Safety-constrained decisions, which require an improvement on the target without harm to a protected outcome, are the hardest decision archetype, and list queries become harder as graphs grow (\cref{app:query}).

\section{Analysis}
\label{sec:analysis}

\subsection{Where \repair investigations stop}
\label{sec:analysis:funnel}

\begin{figure}[t]
    \centering
    \includegraphics[width=0.9\linewidth]{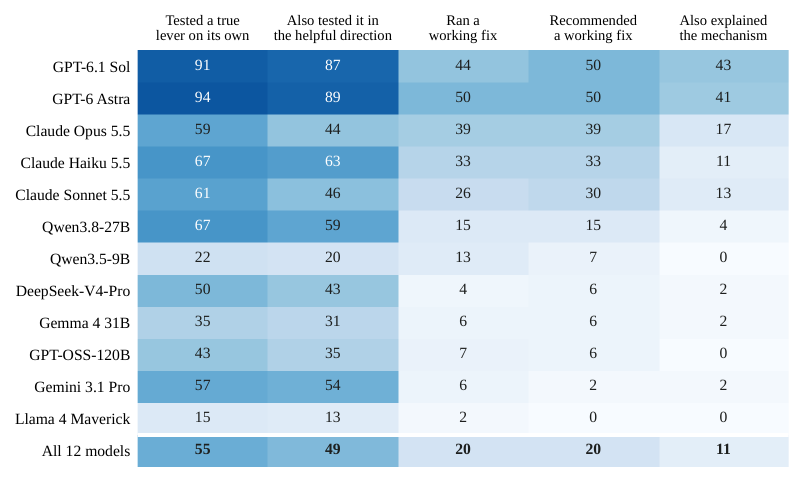}
    \caption{
    Where \repair investigations stop: share of episodes (\%) that reach each milestone, on the evaluation set. Each column is measured independently, and values are from the first run of each model, so they differ from the three-run means in \cref{fig:leaderboard}.
    A working fix is a configuration whose re-simulated benefit is at least 0.9; the last column is acceptance.
    }
    \label{fig:funnel}
\end{figure}

We track five observable milestones of a repair: the agent tests a true lever, tests it in the direction that helps (which the observational data can hide or reverse), runs a working fix, recommends that fix, and explains the mechanism.
Because every executed intervention is a configuration of controls that we can grade as if it were the final recommendation, we record for each episode which of them it reached (\cref{fig:funnel}).
We credit a lever test only when the agent changes that lever on its own, so that its effect can be attributed.

The models stop at different milestones.
Most open-weight models test a true lever on its own in fewer than half of the episodes.
Claude Opus 5.5 and Sonnet 5.5 do it in about 60\% of episodes but test every lever in the helpful direction in only 44--46\%; in confounded and selected worlds the passive correlation can point the wrong way.
The GPT-6 models test the right levers in the helpful direction in about 90\% of episodes, yet only about half of their episodes ever run a working fix; their failed answers most often lack a second control or the right dose.

Once an agent has run a working fix, it recommends it in 95\% of episodes.
Explaining the mechanism is a separate step: only 55\% of the working fixes come with a correct causal account, and the Claude models lose most here.

\subsection{Evidence ledger for \query}
\label{sec:analysis:ledger}

\query cannot be traced against a single passing configuration, so we use an evidence ledger instead.
An LLM annotator marks a turn as \emph{strong evidence} if it computes the correct estimand and the resulting evidence is adequate and supports the gold answer.
For trajectory $i$, let $S_i=1$ if any of its turns is strong evidence and $C_i=1$ if its final answer is correct; we report
\begin{equation}
    \mathrm{Acq} = \Pr(S_i=1), \qquad
    \mathrm{Ret} = \Pr(C_i=1 \mid S_i=1), \qquad
    \mathrm{Ign} = \Pr(S_i=1 \mid C_i=0),
    \label{eq:ledger}
\end{equation}
together with the number of turns spent after the first strong-evidence turn.

In \query, retention is 91\% in modular trajectories without a turn limit, but a third of the failed trajectories had reached strong evidence and still answered wrongly (\cref{app:query_ledger}).
The annotator agrees with a blind human annotator on 85\% of sampled turns ($\kappa=0.69$), and a second annotator model agrees with the human at $\kappa=0.73$.

\begin{figure}[t]
    \centering
    \includegraphics[width=0.85\linewidth]{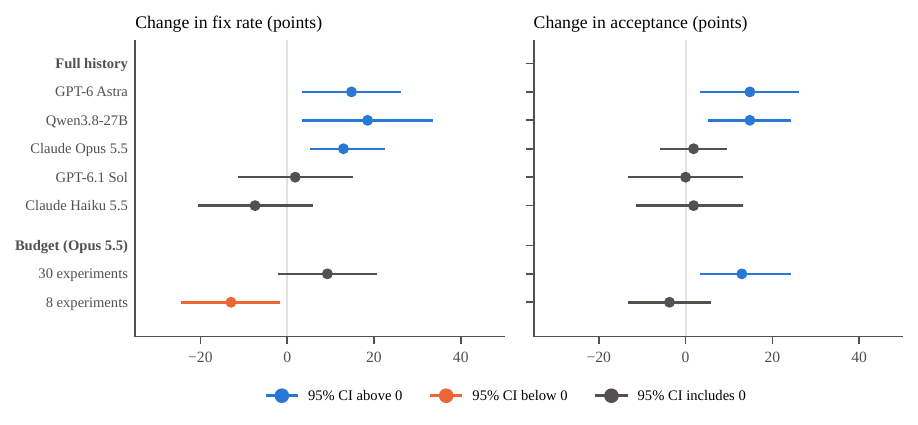}
    \caption{
    Protocol ablations on the evaluation set: paired change in \fixrate{} (left) and acceptance (right) against the model's first run on the same worlds, with 95\% bootstrap intervals over worlds.
    }
    \label{fig:ablations}
\end{figure}

\subsection{Memory across turns}
\label{sec:analysis:memory}

Keeping track of evidence across turns limits some models but not others.
In \repair each turn shows only the last result and the agent's own notes.
Giving agents the full conversation history instead raises GPT-6 Astra's acceptance from 41\% in its first run to 56\% ($p=0.02$), Qwen3.8-27B's from 4\% to 19\% ($p=0.008$), and Claude Opus 5.5's \fixrate{} by 13 points ($p=0.02$), while GPT-6.1 Sol and Claude Haiku 5.5 do not change (\cref{fig:ablations}).
For scale, two runs of GPT-6.1 Sol under the standard protocol differ by 20 points in acceptance (\cref{app:ablations}).
We keep the stateless protocol as the benchmark setting because it gives every model the same information each turn and makes evidence management observable.

\subsection{Budget and resources}
\label{sec:analysis:budget}

The experiment budget matters for Claude Opus 5.5: with 30 experiments its acceptance rises by 13 points ($p=0.04$), and with 8 its \fixrate{} falls by 13 points ($p=0.07$; \cref{fig:ablations}).

In \query, capping sampled rows instead of turns helps the single-prompt active-code agent most.
With at most 5{,}000 sampled rows per task and no turn limit, this agent improves on decision tasks in 8 of 9 model cells (7.3 points on average), more often than the active reasoning and modular agents (\cref{app:query:samp}).

\subsection{Agents and classical procedures}
\label{sec:analysis:scripts}

\begin{table}[t]
\centering
\small
\caption{
Agents and classical procedures with the same budget.
The procedures request controls by identifier and keep all data in memory.
\query: accuracy within the four-turn interface.
\repair: \fixrate{} (\%) on the evaluation set within 15 experiments.
}
\label{tab:baselines}
\begin{tabular}{@{}lcc@{}}
\toprule
\query & \acedcore & \acedqual \\
\midrule
Heuristic policies (random, systematic, estimand, greedy) & 0.07--0.13 & 0.51--0.56 \\
Question-targeted active learning & 0.96 & 1.00 \\
Best \acedcore{} agent (Claude Opus 5.5, modular) & 0.94 & 0.98 \\
\midrule
\repair & \multicolumn{2}{c}{\fixrate{} (\%)} \\
\midrule
Random controls and doses & \multicolumn{2}{c}{13.0} \\
Causal discovery, then tests (best variant on the test set) & \multicolumn{2}{c}{46.3} \\
Rank controls by correlation with the outcome, then tests & \multicolumn{2}{c}{51.9} \\
Best agent (GPT-6 Astra, mean of three runs) & \multicolumn{2}{c}{53.1} \\
\bottomrule
\end{tabular}
\end{table}

Simple classical procedures with the same budget recover working fixes in many single-cause worlds (\cref{tab:baselines}).
On \repair, a procedure that ranks controls by their correlation with the outcome in one measurement batch and then tests them fixes nearly as many worlds as the best agent, and the best causal-discovery pipeline~\citep{spirtes2000causation,chickering2002optimal} followed by the same tests fixes slightly fewer.
Neither procedure states the mechanism well, and neither solves competing-cause or hidden-subtype worlds (\cref{app:baselines}).
On \query, a question-targeted active causal learning policy reaches 0.96 on \acedcore{}, at the level of the best agents.

\section{Limitations and discussion}
\label{sec:limitations}

The scientific control that makes \bench{} evaluable also narrows its scope.
\repair uses continuous SCMs with hidden causes, selection and surrogate outcomes, and \query uses fully observed categorical PGMs with known intervention semantics; neither yet covers temporal dynamics, measurement error, model misspecification, or open-ended experiment design.
The scenarios are synthetic, and code constructs the causal models, enforces intervention semantics, validates margins, and computes oracle answers.
The milestones in \cref{fig:funnel} record which steps an investigation reached; attributing a failure to causal reasoning, statistical inference or search would need further controlled experiments.
A natural next step is \repair worlds in which domain knowledge narrows the search without giving the answer.

The intended positive impact is sharper evaluation of AI systems proposed as scientific assistants.
The main risk is overinterpretation: success in these synthetic hidden-model worlds should not be read as evidence of deployment readiness in real scientific or policy settings, especially where data quality, fairness, privacy, and intervention semantics are contested.

\section{Conclusion}
\label{sec:conclusion}

We introduced \bench{}, a neuro-symbolic benchmark for evaluating LLM agents as active causal investigators in hidden causal worlds.
\repair tests whether agents can diagnose a failing system by choosing experiments and recommend a fix that works on its true objective; \query tests whether they can identify the relevant causal contrast, collect evidence, and return structured causal answers or constrained decisions.
Tracing every \repair experiment shows which milestones investigations reach: models differ in whether they test a true lever, test it in the direction that helps, and find the dose or combination that works.
By combining natural-language scientific scenarios with oracle-scored causal simulators, \bench{} provides a controlled way to measure how future agents acquire and act on causal evidence.

\section*{Acknowledgments}
This work is supported in part by the U.S. Army Research Office under Army-ECASE award W911NF-07-R-0003-03, the U.S. Department of Energy, Office of Science, ARPA-H-SOL-24-101 program, the IARPA HAYSTAC Program, NSF Grants \#2205093, \#2146343, \#2134274, \#2441832, \#2146151 and \#2422214, CDC-RFA-FT-23-0069, and a collaboration with Amazon.
This work used resources of the National Research Platform (NRP), supported in part by NSF awards CNS-1730158, ACI-1540112, ACI-1541349, OAC-1826967, OAC-2112167, CNS-2100237 and CNS-2120019.

\bibliographystyle{unsrtnat}
\bibliography{references}

\newpage
\appendix
\crefalias{section}{appendix}
\crefalias{subsection}{subappendix}

\section{\query details}
\label{app:query}

\begin{figure}[h]
    \centering
    \includegraphics[width=0.9\linewidth]{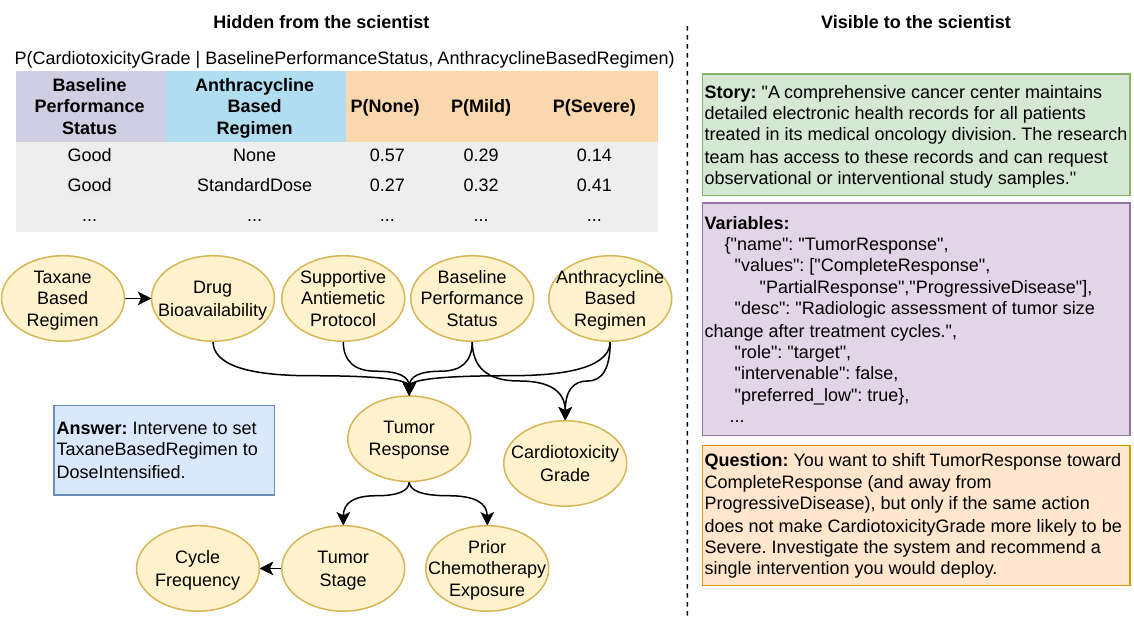}
    \caption{
    Overview of \query on a hard causal-decision task.
    Each task is generated from a hidden PGM.
    The benchmark oracle has access to the causal graph, CPDs, and exact interventional answers.
    The LLM scientist agent only observes the scenario, variable metadata, non-intervenability annotations, and the question, and must answer under a four-turn budget.
    }
    \label{fig:query_overview}
\end{figure}

\begin{figure}[h]
    \centering
    \includegraphics[width=0.9\linewidth]{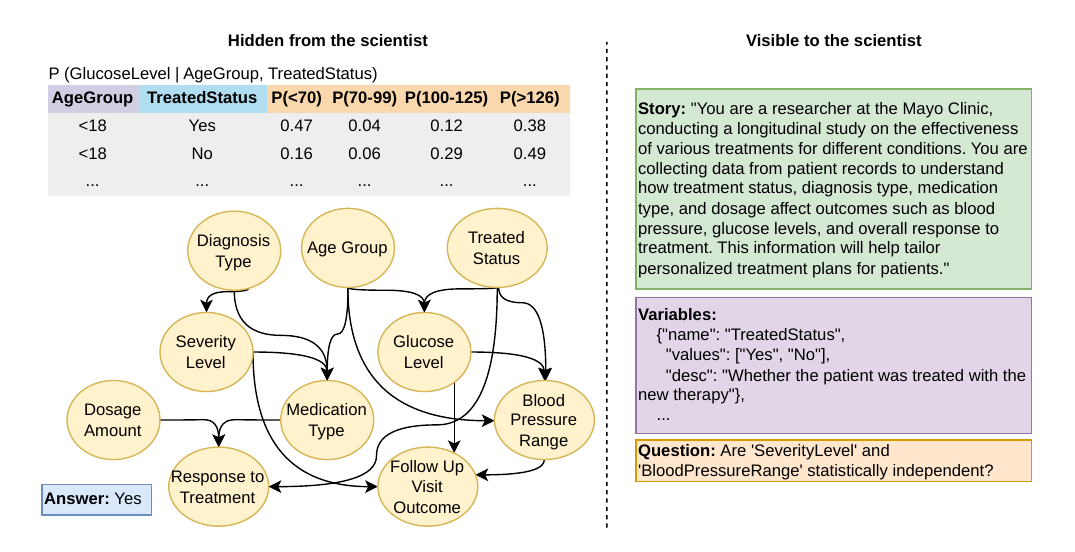}
    \caption{
    The same interface for a structural question: the agent requests observational or interventional samples and answers a yes/no question about the hidden PGM.
    }
    \label{fig:query_structural}
\end{figure}

\begin{figure}[h]
    \centering
    \includegraphics[width=\linewidth]{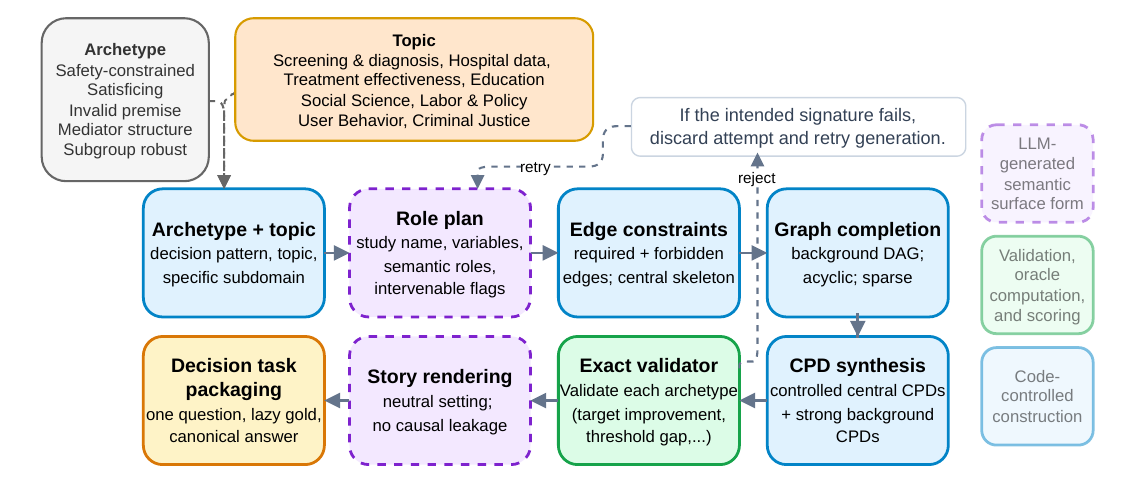}
    \caption{
    Generation pipeline for the causal-decision tasks of \acedcore{}.
    A decision pattern and topic determine semantic roles, graph constraints, CPD construction, validation checks, and the final question.
    LLMs provide readable scientific surface forms, while code enforces causal validity and computes oracle answers.
    }
    \label{fig:query_pipeline}
\end{figure}

\Cref{fig:query_structural,fig:query_pipeline} show the structural interface and how a decision archetype turns a causal signature into a task.
The task types differ in how much experiment selection they require.
\acedqual{} questions give the agent a clear estimand, so active sampling can often estimate the relevant relationship directly.
Causal ancestry/descendancy queries additionally require the agent to plan a set of comparisons, track many candidate variables, and combine the evidence into a set answer, which is why strong agents can nearly solve \acedqual{} while still making errors on the list queries of \acedcore{}.

\query worlds span eight topical domains: Criminal Justice, Education, Hospital Data, Labor \& Policy, Screening \& Diagnosis, Social Science, Treatment Effectiveness, and User Behavior.
Structural worlds have 10, 20, or 30 variables, and causal-decision worlds have 10 or 15.
For structural worlds, question labels are determined by graph ancestry and d-separation, and CPDs are resampled until every dependence the graph implies exceeds a total-variation margin of 0.02.
For causal-decision worlds, each archetype has an exact-inference validator that requires a clear oracle answer, for example a safe action that improves the target by at least 0.20 expected-state-index units while changing the safety outcome by at most 0.08.

Every task scores 0 or 1, and \acedcore{} weights all tasks equally.
An LLM extracts the final answer from the agent's reply (yes or no, a list of variables, or a (variable, value) intervention), which is then scored deterministically: list answers are matched to the variable catalogue, and decisions are compared with the oracle effect table.
Exact matching rewards cautious list answers: 7 of the 10 \acedcore{} list queries have an empty answer set.
Intervention effects are compared through the expected-state index of each outcome's ordered states, with ties accepted within 0.05.
A safe action may change the safety outcome by at most 0.05, and a subgroup-robust action must improve each subgroup by at least 0.15.
Mediator and invalid-premise tasks are scored by their categorical label.
One subgroup-robust task cannot be answered correctly, because its oracle action improves one subgroup by only 0.145.
Pandas reads a state named ``None'' as missing, so the agents' data summaries omit that state in the 66 of 70 worlds that have one.

\Cref{tab:query_full} gives every model and agent with intervals.
\Cref{tab:query_archetype} breaks the decisions down by archetype, and \cref{tab:query_pooled} pools all models by world property: safety-constrained decisions are the hardest archetype, and list queries become harder with graph size and edge density.

\begin{table}[t]
\centering
\footnotesize
\setlength{\tabcolsep}{3.5pt}
\caption{\query accuracy for every model and agent, with 95\% world-clustered bootstrap intervals. Rows are in the order of \cref{tab:query}; two Active cells, Claude Sonnet 5.5 and Qwen3.6-27B, have 69 \acedcore{} tasks.}
\label{tab:query_full}
\begin{tabular}{@{}lcccc@{}}
\toprule
Model & Guess-only & Active & Active+Code & Modular \\
\midrule
\multicolumn{5}{@{}l}{\textit{\acedcore}} \\
Claude Sonnet 5.5 & 0.40\,{\scriptsize[0.28, 0.51]} & 0.86\,{\scriptsize[0.77, 0.94]} & 0.89\,{\scriptsize[0.81, 0.96]} & 0.91\,{\scriptsize[0.84, 0.97]} \\
Claude Opus 5.5 & 0.49\,{\scriptsize[0.37, 0.60]} & 0.83\,{\scriptsize[0.74, 0.91]} & 0.86\,{\scriptsize[0.77, 0.94]} & 0.94\,{\scriptsize[0.89, 0.99]} \\
Gemini 3.1 Pro & 0.58\,{\scriptsize[0.47, 0.69]} & 0.83\,{\scriptsize[0.74, 0.91]} & 0.79\,{\scriptsize[0.69, 0.87]} & 0.90\,{\scriptsize[0.83, 0.97]} \\
Claude Haiku 5.5 & 0.47\,{\scriptsize[0.36, 0.59]} & 0.76\,{\scriptsize[0.66, 0.86]} & 0.73\,{\scriptsize[0.61, 0.83]} & 0.86\,{\scriptsize[0.77, 0.93]} \\
Gemma 4 31B & 0.45\,{\scriptsize[0.35, 0.56]} & 0.73\,{\scriptsize[0.63, 0.83]} & 0.77\,{\scriptsize[0.67, 0.87]} & 0.79\,{\scriptsize[0.69, 0.89]} \\
Qwen3.8-27B & 0.41\,{\scriptsize[0.29, 0.53]} & 0.77\,{\scriptsize[0.67, 0.87]} & 0.74\,{\scriptsize[0.64, 0.84]} & 0.76\,{\scriptsize[0.66, 0.86]} \\
Claude Opus 4.6 & 0.41\,{\scriptsize[0.31, 0.53]} & 0.76\,{\scriptsize[0.66, 0.86]} & 0.77\,{\scriptsize[0.67, 0.87]} & 0.71\,{\scriptsize[0.61, 0.81]} \\
DeepSeek-V4-Pro & 0.42\,{\scriptsize[0.32, 0.54]} & 0.71\,{\scriptsize[0.61, 0.81]} & 0.81\,{\scriptsize[0.71, 0.90]} & 0.64\,{\scriptsize[0.53, 0.76]} \\
Qwen3.6-27B & 0.44\,{\scriptsize[0.32, 0.55]} & 0.71\,{\scriptsize[0.61, 0.81]} & 0.70\,{\scriptsize[0.59, 0.80]} & 0.76\,{\scriptsize[0.66, 0.86]} \\
Claude Sonnet 4.6 & 0.53\,{\scriptsize[0.41, 0.64]} & 0.80\,{\scriptsize[0.71, 0.89]} & 0.61\,{\scriptsize[0.50, 0.73]} & 0.69\,{\scriptsize[0.57, 0.79]} \\
Llama 4 Maverick & 0.41\,{\scriptsize[0.31, 0.53]} & 0.66\,{\scriptsize[0.54, 0.76]} & 0.61\,{\scriptsize[0.50, 0.73]} & 0.67\,{\scriptsize[0.56, 0.77]} \\
GPT-6.1 Sol & 0.42\,{\scriptsize[0.31, 0.54]} & 0.63\,{\scriptsize[0.51, 0.74]} & 0.77\,{\scriptsize[0.67, 0.87]} & 0.47\,{\scriptsize[0.36, 0.59]} \\
GPT-6 Astra & 0.45\,{\scriptsize[0.33, 0.56]} & 0.64\,{\scriptsize[0.53, 0.76]} & 0.77\,{\scriptsize[0.67, 0.87]} & 0.41\,{\scriptsize[0.30, 0.53]} \\
Qwen3.5-9B & 0.46\,{\scriptsize[0.36, 0.58]} & 0.63\,{\scriptsize[0.51, 0.74]} & 0.60\,{\scriptsize[0.49, 0.71]} & 0.54\,{\scriptsize[0.43, 0.66]} \\
Llama 3.3 70B & 0.38\,{\scriptsize[0.28, 0.50]} & 0.64\,{\scriptsize[0.53, 0.76]} & 0.37\,{\scriptsize[0.26, 0.49]} & 0.60\,{\scriptsize[0.49, 0.71]} \\
GPT-OSS-120B & 0.45\,{\scriptsize[0.35, 0.56]} & 0.53\,{\scriptsize[0.41, 0.64]} & 0.19\,{\scriptsize[0.10, 0.27]} & 0.51\,{\scriptsize[0.40, 0.63]} \\
GPT-4o & 0.40\,{\scriptsize[0.29, 0.51]} & 0.44\,{\scriptsize[0.31, 0.56]} & 0.33\,{\scriptsize[0.23, 0.44]} & 0.39\,{\scriptsize[0.27, 0.50]} \\
\midrule
\multicolumn{5}{@{}l}{\textit{\acedqual}} \\
Claude Sonnet 5.5 & 0.55\,{\scriptsize[0.49, 0.60]} & 1.00\,{\scriptsize[1.00, 1.00]} & 1.00\,{\scriptsize[1.00, 1.00]} & 1.00\,{\scriptsize[1.00, 1.00]} \\
Claude Opus 5.5 & 0.53\,{\scriptsize[0.48, 0.59]} & 0.98\,{\scriptsize[0.94, 1.00]} & 0.98\,{\scriptsize[0.94, 1.00]} & 0.98\,{\scriptsize[0.94, 1.00]} \\
Gemini 3.1 Pro & 0.47\,{\scriptsize[0.42, 0.53]} & 0.98\,{\scriptsize[0.94, 1.00]} & 1.00\,{\scriptsize[1.00, 1.00]} & 0.96\,{\scriptsize[0.90, 1.00]} \\
Claude Haiku 5.5 & 0.48\,{\scriptsize[0.42, 0.53]} & 0.90\,{\scriptsize[0.80, 0.98]} & 0.98\,{\scriptsize[0.94, 1.00]} & 1.00\,{\scriptsize[1.00, 1.00]} \\
Gemma 4 31B & 0.47\,{\scriptsize[0.40, 0.53]} & 0.86\,{\scriptsize[0.74, 0.98]} & 0.80\,{\scriptsize[0.74, 0.86]} & 0.94\,{\scriptsize[0.88, 1.00]} \\
Qwen3.8-27B & 0.53\,{\scriptsize[0.49, 0.58]} & 0.98\,{\scriptsize[0.94, 1.00]} & 0.98\,{\scriptsize[0.94, 1.00]} & 0.92\,{\scriptsize[0.86, 0.98]} \\
Claude Opus 4.6 & 0.53\,{\scriptsize[0.49, 0.56]} & 1.00\,{\scriptsize[1.00, 1.00]} & 1.00\,{\scriptsize[1.00, 1.00]} & 0.64\,{\scriptsize[0.56, 0.74]} \\
DeepSeek-V4-Pro & 0.49\,{\scriptsize[0.45, 0.54]} & 0.92\,{\scriptsize[0.86, 0.98]} & 0.96\,{\scriptsize[0.90, 1.00]} & 0.82\,{\scriptsize[0.70, 0.92]} \\
Qwen3.6-27B & 0.49\,{\scriptsize[0.44, 0.55]} & 0.90\,{\scriptsize[0.84, 0.96]} & 0.94\,{\scriptsize[0.88, 1.00]} & 0.98\,{\scriptsize[0.94, 1.00]} \\
Claude Sonnet 4.6 & 0.49\,{\scriptsize[0.44, 0.54]} & 0.98\,{\scriptsize[0.94, 1.00]} & 0.76\,{\scriptsize[0.68, 0.84]} & 0.90\,{\scriptsize[0.84, 0.96]} \\
Llama 4 Maverick & 0.51\,{\scriptsize[0.44, 0.57]} & 0.78\,{\scriptsize[0.70, 0.88]} & 0.78\,{\scriptsize[0.68, 0.88]} & 0.88\,{\scriptsize[0.80, 0.96]} \\
GPT-6.1 Sol & 0.54\,{\scriptsize[0.49, 0.59]} & 0.76\,{\scriptsize[0.70, 0.80]} & 0.92\,{\scriptsize[0.86, 0.98]} & 0.66\,{\scriptsize[0.54, 0.78]} \\
GPT-6 Astra & 0.53\,{\scriptsize[0.48, 0.57]} & 0.76\,{\scriptsize[0.70, 0.80]} & 0.94\,{\scriptsize[0.84, 1.00]} & 0.62\,{\scriptsize[0.50, 0.72]} \\
Qwen3.5-9B & 0.50\,{\scriptsize[0.43, 0.57]} & 0.74\,{\scriptsize[0.54, 0.90]} & 0.56\,{\scriptsize[0.44, 0.66]} & 0.74\,{\scriptsize[0.64, 0.84]} \\
Llama 3.3 70B & 0.51\,{\scriptsize[0.47, 0.56]} & 0.72\,{\scriptsize[0.56, 0.84]} & 0.66\,{\scriptsize[0.56, 0.76]} & 0.88\,{\scriptsize[0.76, 0.98]} \\
GPT-OSS-120B & 0.48\,{\scriptsize[0.43, 0.53]} & 0.84\,{\scriptsize[0.72, 0.96]} & 0.34\,{\scriptsize[0.18, 0.50]} & 0.70\,{\scriptsize[0.58, 0.82]} \\
GPT-4o & 0.49\,{\scriptsize[0.41, 0.57]} & 0.84\,{\scriptsize[0.72, 0.94]} & 0.42\,{\scriptsize[0.32, 0.52]} & 0.74\,{\scriptsize[0.62, 0.84]} \\
\bottomrule
\end{tabular}
\end{table}

\begin{table}[t]
\centering
\footnotesize
\setlength{\tabcolsep}{2.6pt}
\caption{Causal-decision accuracy (\%) by archetype, model and agent on \acedcore{}. IP: invalid premise; Med: mediator structure; Saf: safety-constrained; Sat: satisficing; Sub: subgroup-robust. The last row pools all 17 models.}
\label{tab:query_archetype}
\begin{tabular}{@{}l*{5}{c}@{\hspace{6pt}}*{5}{c}@{\hspace{6pt}}*{5}{c}@{}}
\toprule
 & \multicolumn{5}{c}{Active} & \multicolumn{5}{c}{Active+Code} & \multicolumn{5}{c}{Modular} \\
\cmidrule(lr){2-6}\cmidrule(lr){7-11}\cmidrule(l){12-16}
Model & IP & Med & Saf & Sat & Sub & IP & Med & Saf & Sat & Sub & IP & Med & Saf & Sat & Sub \\
\midrule
Claude Sonnet 5.5 & 92 & 100 & 73 & 92 & 83 & 100 & 100 & 58 & 100 & 92 & 100 & 100 & 67 & 100 & 92 \\
Claude Opus 5.5 & 83 & 92 & 67 & 92 & 92 & 100 & 92 & 92 & 75 & 92 & 100 & 100 & 83 & 100 & 92 \\
Gemini 3.1 Pro & 83 & 83 & 83 & 92 & 92 & 75 & 100 & 33 & 100 & 92 & 83 & 100 & 75 & 100 & 92 \\
Claude Haiku 5.5 & 83 & 100 & 83 & 67 & 83 & 83 & 67 & 58 & 75 & 83 & 100 & 83 & 75 & 92 & 75 \\
Gemma 4 31B & 75 & 75 & 67 & 92 & 92 & 83 & 67 & 67 & 92 & 92 & 92 & 75 & 67 & 92 & 92 \\
Qwen3.8-27B & 75 & 83 & 58 & 92 & 92 & 83 & 92 & 67 & 83 & 67 & 75 & 100 & 33 & 75 & 83 \\
Claude Opus 4.6 & 92 & 83 & 67 & 92 & 92 & 75 & 100 & 42 & 92 & 83 & 67 & 67 & 67 & 92 & 92 \\
DeepSeek-V4-Pro & 75 & 83 & 50 & 92 & 92 & 100 & 92 & 58 & 100 & 92 & 83 & 75 & 42 & 75 & 75 \\
Qwen3.6-27B & 75 & 73 & 42 & 92 & 92 & 83 & 58 & 42 & 92 & 83 & 83 & 100 & 42 & 75 & 83 \\
Claude Sonnet 4.6 & 92 & 92 & 67 & 83 & 83 & 58 & 75 & 50 & 75 & 67 & 75 & 92 & 67 & 58 & 75 \\
Llama 4 Maverick & 75 & 83 & 67 & 75 & 67 & 75 & 50 & 58 & 75 & 83 & 92 & 42 & 75 & 58 & 92 \\
GPT-6.1 Sol & 75 & 58 & 33 & 92 & 83 & 92 & 100 & 25 & 100 & 92 & 75 & 42 & 0 & 67 & 75 \\
GPT-6 Astra & 75 & 33 & 58 & 92 & 83 & 83 & 83 & 50 & 100 & 92 & 75 & 58 & 0 & 25 & 75 \\
Qwen3.5-9B & 75 & 75 & 17 & 83 & 83 & 75 & 42 & 42 & 75 & 83 & 50 & 33 & 58 & 67 & 50 \\
Llama 3.3 70B & 67 & 75 & 50 & 83 & 75 & 8 & 50 & 17 & 67 & 58 & 75 & 58 & 75 & 50 & 67 \\
GPT-OSS-120B & 83 & 42 & 42 & 92 & 33 & 17 & 25 & 8 & 17 & 25 & 67 & 50 & 42 & 33 & 75 \\
GPT-4o & 50 & 67 & 8 & 67 & 58 & 25 & 25 & 0 & 83 & 58 & 50 & 33 & 0 & 50 & 83 \\
\midrule
All models & 78 & 76 & 55 & 86 & 81 & 72 & 72 & 45 & 82 & 78 & 79 & 71 & 51 & 71 & 80 \\
\bottomrule
\end{tabular}
\end{table}

\begin{table}[t]
\centering
\small
\setlength{\tabcolsep}{3.5pt}
\caption{\query accuracy by world property, pooled over the 17 models. A: Active; A+C: Active+Code; M: Modular. Decision worlds have 10 or 15 variables and at most nine intervenable variables; structural worlds have 10, 20, or 30 variables.}
\label{tab:query_pooled}
\begin{tabular}{@{}lccc@{\hspace{8pt}}ccc@{\hspace{8pt}}ccc@{}}
\toprule
 & \multicolumn{3}{c}{\acedqual} & \multicolumn{3}{c}{\acedcore{} lists} & \multicolumn{3}{c}{\acedcore{} decisions} \\
\cmidrule(lr){2-4}\cmidrule(lr){5-7}\cmidrule(l){8-10}
 & A & A+C & M & A & A+C & M & A & A+C & M \\
\midrule
\multicolumn{10}{@{}l}{\textit{Variables}} \\
\quad 10 & 0.87 & 0.83 & 0.84 & 0.63 & 0.54 & 0.66 & 0.73 & 0.69 & 0.68 \\
\quad 15 & -- & -- & -- & -- & -- & -- & 0.77 & 0.71 & 0.73 \\
\quad 20 & 0.87 & 0.80 & 0.85 & 0.27 & 0.53 & 0.51 & -- & -- & -- \\
\quad 30 & 0.90 & 0.84 & 0.85 & 0.20 & 0.31 & 0.37 & -- & -- & -- \\
\midrule
\multicolumn{10}{@{}l}{\textit{Edges per variable}} \\
\quad $\le$1.0 & 0.85 & 0.82 & 0.83 & 0.78 & 0.71 & 0.75 & 0.78 & 0.73 & 0.73 \\
\quad 1.0--1.5 & 0.89 & 0.83 & 0.85 & 0.23 & 0.37 & 0.44 & 0.49 & 0.44 & 0.48 \\
\midrule
\multicolumn{10}{@{}l}{\textit{Intervenable variables}} \\
\quad $\le$9 & 0.82 & 0.83 & 0.82 & 0.47 & 0.41 & 0.56 & 0.75 & 0.70 & 0.70 \\
\quad 10--14 & 0.88 & 0.81 & 0.84 & 0.61 & 0.61 & 0.69 & -- & -- & -- \\
\quad 15--20 & 0.90 & 0.83 & 0.87 & 0.19 & 0.43 & 0.44 & -- & -- & -- \\
\quad $>$20 & 0.91 & 0.84 & 0.82 & 0.41 & 0.35 & 0.35 & -- & -- & -- \\
\midrule
\multicolumn{10}{@{}l}{\textit{Domain}} \\
\quad Criminal Justice & 0.91 & 0.83 & 0.86 & 0.41 & 0.38 & 0.50 & 0.66 & 0.65 & 0.68 \\
\quad Education & 0.86 & 0.82 & 0.85 & 0.15 & 0.47 & 0.50 & 0.89 & 0.75 & 0.78 \\
\quad Hospital Data & 0.89 & 0.82 & 0.82 & 0.88 & 0.71 & 0.82 & 0.71 & 0.71 & 0.73 \\
\quad Labor \& Policy & 0.94 & 0.78 & 0.85 & 0.12 & 0.35 & 0.29 & 0.71 & 0.64 & 0.64 \\
\quad Screening \& Diagnosis & 0.71 & 0.80 & 0.78 & 0.76 & 0.76 & 0.71 & 0.84 & 0.69 & 0.67 \\
\quad Social Science & 0.91 & 0.84 & 0.82 & 0.41 & 0.35 & 0.35 & 0.98 & 0.84 & 0.69 \\
\quad Treatment Effectiveness & 0.87 & 0.86 & 0.88 & 0.47 & 0.76 & 0.71 & 0.48 & 0.53 & 0.60 \\
\quad User Behavior & 0.93 & 0.86 & 0.86 & 0.18 & 0.06 & 0.41 & 0.97 & 0.94 & 0.85 \\
\bottomrule
\end{tabular}
\end{table}

\subsection{Sampled-row budget}
\label{app:query:samp}

\begin{figure}[h]
    \centering
    \includegraphics[width=0.9\linewidth]{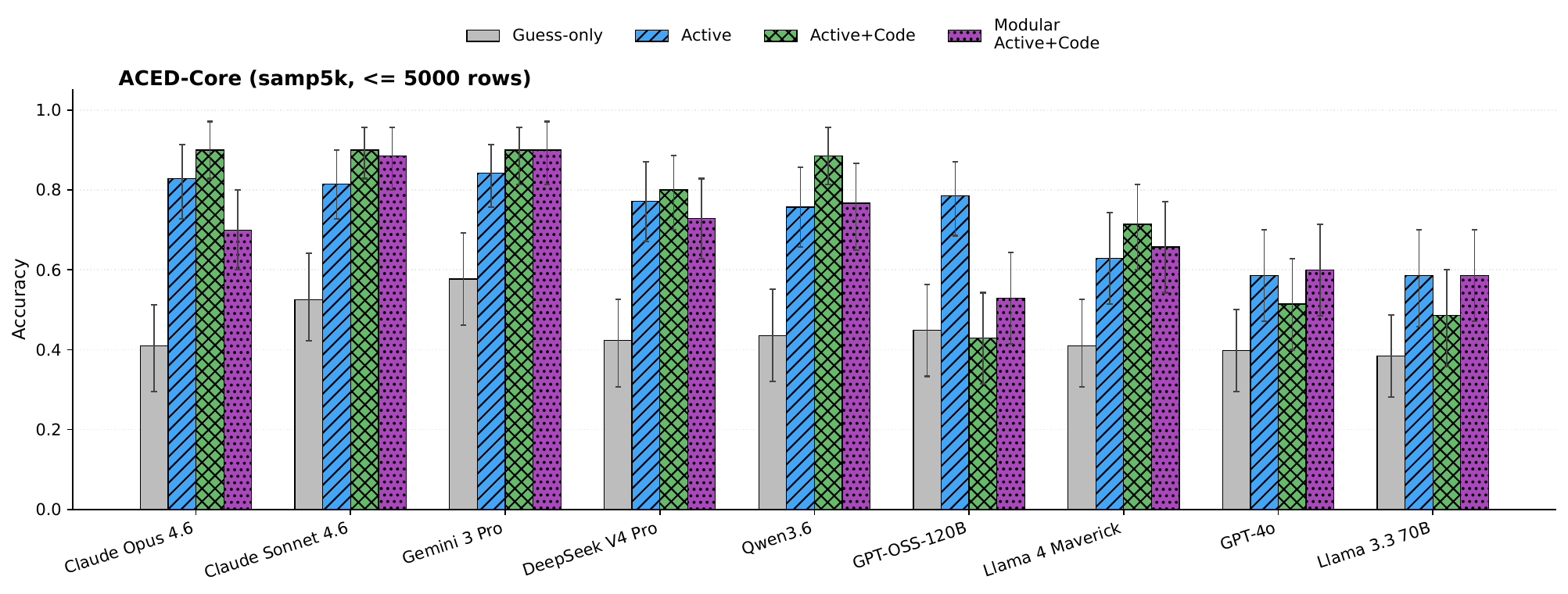}
    \caption{\acedcore{} accuracy when each task may use at most 5{,}000 sampled rows instead of four turns.}
    \label{fig:query_samp}
\end{figure}

The main setting caps the turns.
In a sample-budget ablation, each task may instead use at most 5{,}000 sampled rows (and at most ten data queries), so an agent can still iterate but must allocate a limited number of samples across candidate contrasts (\cref{fig:query_samp}).
On paired causal-decision cells of the nine models run in both settings, the single-prompt active-code agent improves in 8 of 9 cells (7.3 points on average), the active reasoning agent in 6 of 9 (4.2 points), and the modular agent in 4 of 9 (2.6 points).
We do not use the sampled-row setting for the leaderboard.

\section{\query evidence ledger}
\label{app:query_ledger}

\begin{figure}[t]
    \centering
    \includegraphics[width=0.8\linewidth]{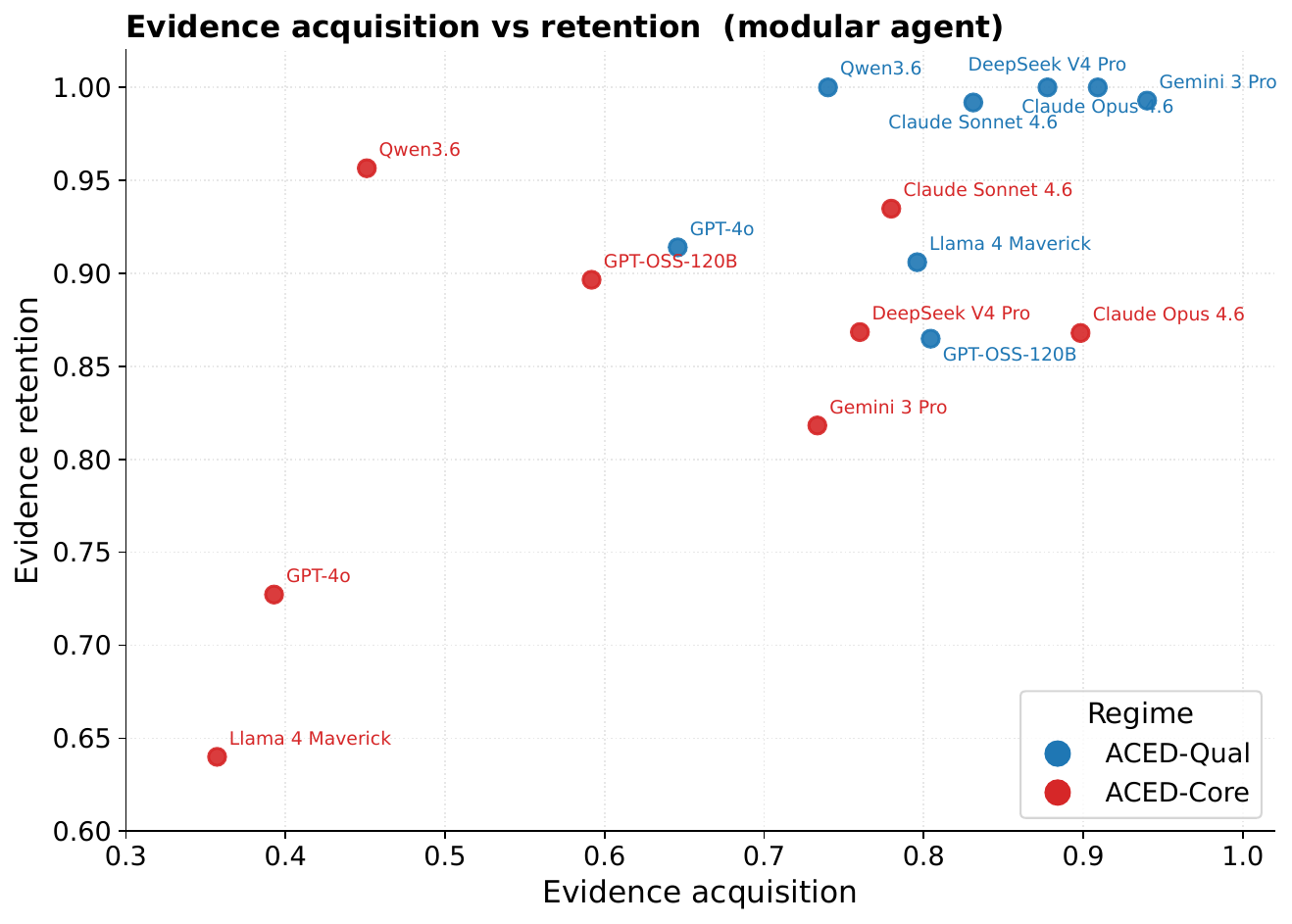}
    \caption{
    Evidence acquisition versus retention for modular active-code trajectories on \acedqual{} and \acedcore{}; bubble size indicates ignored evidence.
    }
    \label{fig:query_ledger_scatter}
\end{figure}

\begin{figure}[t]
    \centering
    \includegraphics[width=\linewidth]{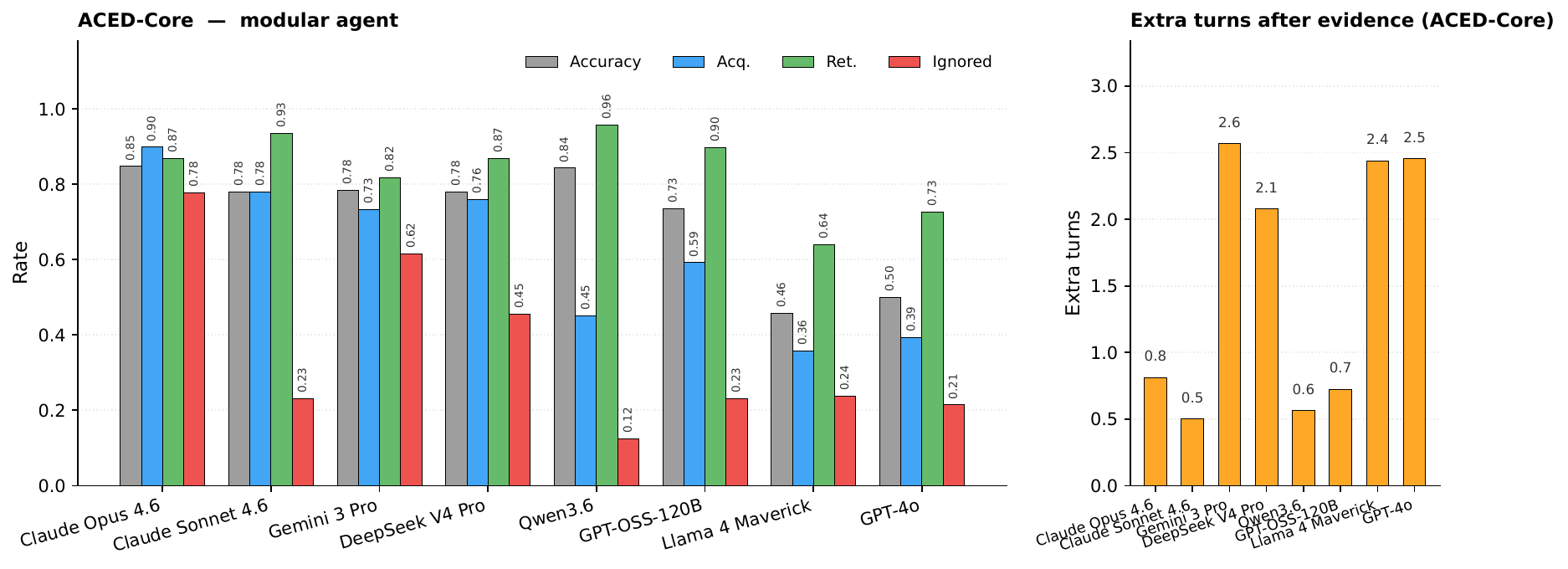}
    \caption{
    Evidence-ledger diagnostics for Modular active-code trajectories on \acedcore{}, pooled over three budget settings.
    The left panel reports final-answer accuracy, evidence acquisition, evidence retention, and ignored evidence.
    The right panel reports extra turns after the first strong-evidence turn.
    }
    \label{fig:query_ledger_bars}
\end{figure}

For each turn of a modular active-code trajectory, an LLM annotator reads the turn's query, code, and output and judges whether the turn computed the correct estimand, whether the resulting evidence supports the gold answer, and whether that evidence is adequate; a turn is \emph{strong evidence} only if all three hold.
A trajectory has $S_i=1$ if any of its turns is strong evidence, and the ledger metrics follow \cref{eq:ledger} (\cref{fig:query_ledger_bars,fig:query_ledger_scatter}).
The ledger covers 3{,}579 modular trajectories from the four-turn setting, the sampled-row setting, and a setting without a turn limit.
In the setting without a turn limit, retention is 91\% and 35\% of failed trajectories had reached strong evidence.

We validate the turn labels on 48 anonymised turns, labelled by a blind human annotator and by GPT-4o as a second model annotator (\cref{tab:query_kappa}).
All three annotators mark a similar share of turns as valid evidence.

\begin{figure}[t]
    \centering
    \includegraphics[width=\linewidth]{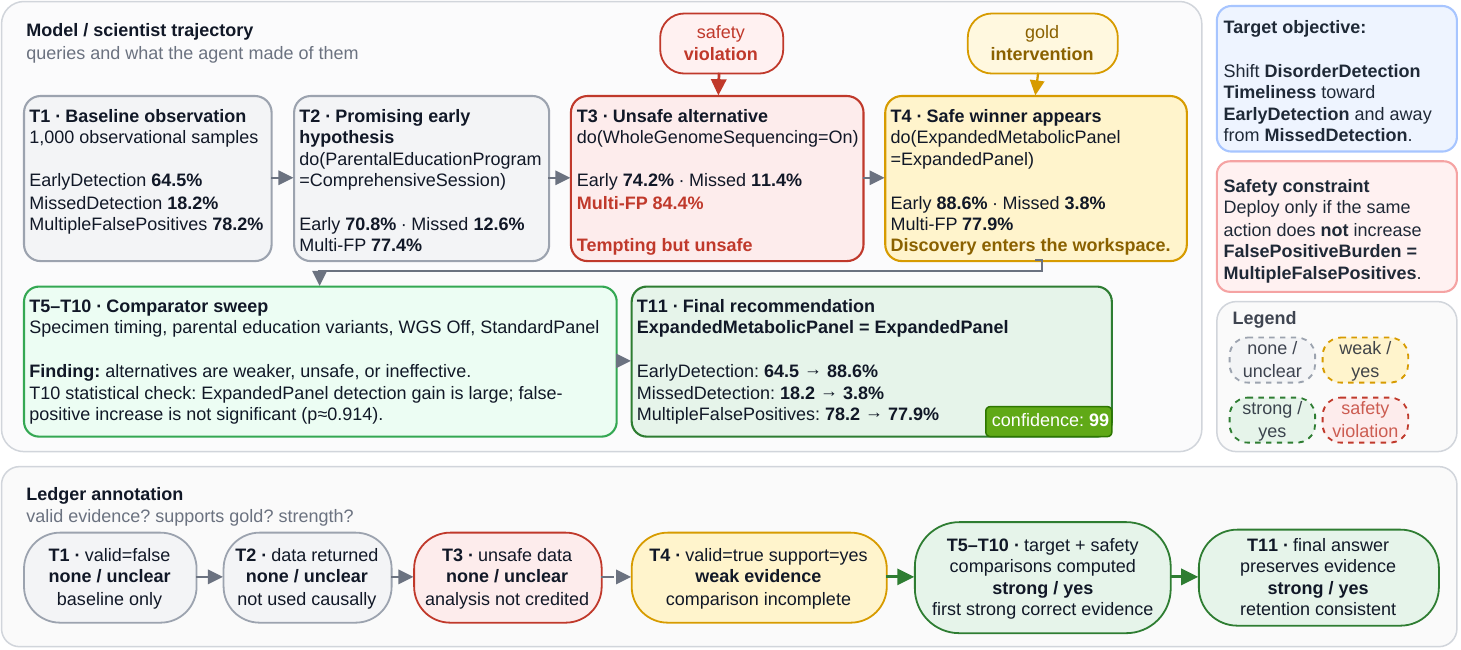}
    \caption{
    Evidence-ledger trace for a safety-constrained \acedcore{} decision task.
    Strong evidence appears only after the agent compares candidate interventions on both the target and the safety outcome.
    }
    \label{fig:query_trace}
\end{figure}

\Cref{fig:query_trace} shows a safety-constrained decision in which the agent must improve \texttt{DisorderDetectionTimeliness} without increasing \texttt{FalsePositiveBurden}.
The early turns are insufficient because the agent has not yet compared candidate interventions on both outcomes.
The decisive comparison supports \texttt{ExpandedMetabolicPanel=ExpandedPanel}: early detection rises from 64.5\% to 88.6\%, while the safety outcome stays essentially unchanged, and the final answer keeps this recommendation.
In a subgroup-robust task, the ledger marks strong evidence only once the agent compares interventions within each subgroup, because the action with the largest effect in one subgroup is weaker in the other.

\begin{table}[t]
\centering
\small
\caption{Agreement between ledger annotators on 48 anonymised turns: raw agreement and Cohen's $\kappa$ for the valid-evidence and supports-gold labels.}
\label{tab:query_kappa}
\begin{tabular}{@{}lcccc@{}}
\toprule
 & \multicolumn{2}{c}{Valid evidence} & \multicolumn{2}{c}{Supports gold} \\
\cmidrule(lr){2-3}\cmidrule(l){4-5}
Pair & Agreement & $\kappa$ & Agreement & $\kappa$ \\
\midrule
Human vs.\ Claude annotator & 85.4\% & 0.692 & 85.4\% & 0.721 \\
Human vs.\ GPT-4o & 87.5\% & 0.734 & 83.3\% & 0.675 \\
Claude annotator vs.\ GPT-4o & 81.2\% & 0.596 & 85.4\% & 0.701 \\
\midrule
\multicolumn{5}{@{}l}{Turns labelled valid evidence: human 60.4\%, Claude annotator 62.5\%, GPT-4o 64.6\%} \\
\bottomrule
\end{tabular}
\end{table}

\section{Classical procedures and leakage audit}
\label{app:query_baselines}

On \query, the reference policies use the agents' interface: at most three data queries and an answer per question, through the same simulator and scorer.
The question-targeted policy spends its queries on the estimand the question names and answers decisions by successive elimination over candidate interventions; the structure-first policies learn an equivalence class with PC or GES and orient it with interventions chosen by the criterion of \citet{he2008active}.
Four heuristic policies (random, systematic, estimand-informed, association-greedy) were written for structural questions and return fixed recommendations on decisions.
\Cref{tab:query_acl} reports all of them with five seeds.

\begin{table}[t]
\centering
\footnotesize
\setlength{\tabcolsep}{3.5pt}
\caption{Reference policies on \query, on the tasks of the active agents (four-turn interface, five seeds). Rows: sample rows per query. Lists and decisions are the two parts of \acedcore{}. 95\% world-clustered intervals.}
\label{tab:query_acl}
\begin{tabular}{@{}lrcccc@{}}
\toprule
Policy & Rows & \acedqual & Lists & Decisions & \acedcore \\
\midrule
Question-targeted + elimination & 10{,}000 & 1.00\,{\scriptsize[0.99, 1.00]} & 0.86 & 0.98 & 0.96\,{\scriptsize[0.92, 1.00]} \\
 & 2{,}000 & 0.99\,{\scriptsize[0.98, 1.00]} & 0.86 & 0.95 & 0.94\,{\scriptsize[0.89, 0.98]} \\
 & 500 & 0.97\,{\scriptsize[0.94, 0.99]} & 0.66 & 0.91 & 0.87\,{\scriptsize[0.81, 0.93]} \\
ACL-PC + elimination & 10{,}000 & 0.93\,{\scriptsize[0.88, 0.98]} & 0.64 & 0.98 & 0.93\,{\scriptsize[0.87, 0.98]} \\
 & 2{,}000 & 0.92\,{\scriptsize[0.86, 0.98]} & 0.44 & 0.95 & 0.88\,{\scriptsize[0.81, 0.94]} \\
 & 500 & 0.82\,{\scriptsize[0.74, 0.90]} & 0.48 & 0.91 & 0.85\,{\scriptsize[0.78, 0.91]} \\
ACL-GES + elimination & 10{,}000 & 0.92\,{\scriptsize[0.87, 0.98]} & 0.60 & 0.98 & 0.93\,{\scriptsize[0.87, 0.97]} \\
 & 2{,}000 & 0.92\,{\scriptsize[0.88, 0.97]} & 0.44 & 0.95 & 0.88\,{\scriptsize[0.80, 0.95]} \\
 & 500 & 0.75\,{\scriptsize[0.64, 0.86]} & 0.56 & 0.91 & 0.86\,{\scriptsize[0.79, 0.92]} \\
\midrule
Random & 1{,}000 & 0.51\,{\scriptsize[0.48, 0.54]} & 0.12 & 0.13 & 0.13\,{\scriptsize[0.09, 0.17]} \\
Systematic & 1{,}000 & 0.56\,{\scriptsize[0.50, 0.60]} & 0.48 & 0.05 & 0.11\,{\scriptsize[0.04, 0.19]} \\
Estimand-informed & 1{,}000 & 0.54\,{\scriptsize[0.42, 0.64]} & 0.24 & 0.05 & 0.08\,{\scriptsize[0.03, 0.14]} \\
Association-greedy & 1{,}000 & 0.55\,{\scriptsize[0.50, 0.60]} & 0.18 & 0.05 & 0.07\,{\scriptsize[0.01, 0.13]} \\
\midrule
Best agent (Claude Opus 5.5, Modular) & -- & 0.98\,{\scriptsize[0.94, 1.00]} & -- & -- & 0.94\,{\scriptsize[0.89, 0.99]} \\
\bottomrule
\end{tabular}
\end{table}

To test whether names leak answers, we anonymise 60 structural worlds (variables become $X_1, \dots, X_n$, states $s_0, \dots, s_k$, and the scenario a generic description) and compare guess-only accuracy on the same 360 questions with the same model, prompt, output limit and extractor (\cref{tab:query_leakage}).
We find no evidence that surface form leaks answers.

\begin{table}[t]
\centering
\small
\caption{Matched leakage audit: guess-only accuracy on 360 structural questions with named and anonymised worlds. Same model, prompt, output limit and extractor in both conditions; 95\% world-bootstrap intervals; exact McNemar test.}
\label{tab:query_leakage}
\begin{tabular}{@{}lcccc@{}}
\toprule
Model & Named & Anonymised & Difference & $p$ \\
\midrule
Claude Opus 5.5 & 45.3 {\scriptsize[41.9, 48.9]} & 44.4 {\scriptsize[40.6, 48.3]} & $-0.8$ {\scriptsize[$-5.6$, $+3.9$]} & 0.84 \\
Claude Haiku 5.5 & 45.0 {\scriptsize[41.4, 48.6]} & 43.6 {\scriptsize[40.0, 47.2]} & $-1.4$ {\scriptsize[$-5.8$, $+3.1$]} & 0.68 \\
\bottomrule
\end{tabular}
\end{table}

\subsection{\repair procedures}
\label{app:baselines}

The \repair procedures run through the same simulator and grader as the agents, request controls by catalogue identifier, and keep every row in memory.
\emph{Random} sets one to three random controls to random doses in each experiment.
The \emph{correlation-ranking} procedure takes one observational batch, ranks every setting by the correlation of its level with the outcome, tests the top three at interior values in both directions, and searches the dose of the best one or two.
The \emph{causal-discovery} procedures learn a graph from an observational batch or from a do-nothing interventional batch with PC, FCI or GES~\citep{spirtes2000causation,chickering2002optimal} and test the settings adjacent to the outcome in the same way; the best variant uses PC on the interventional batch.
On all 540 test worlds they fix 8.3\%, 52.8\% and 40--45\% of worlds, and neither the correlation-ranking procedure nor the best causal-discovery procedure fixes any competing-cause or hidden-subtype world.

\section{\repair worlds and scoring}
\label{app:repair}

Each world combines one of the nine causal structures (\cref{tab:structures}) with one of ten domains: greenhouse agronomy, aquaculture, battery manufacturing, bioprocessing, catalysis, clinical care, data centres, fermentation, semiconductor fabrication and water treatment.
The structure fixes the graph around the hidden cause, and the domain supplies realistic names, units, ranges and usual levels.
Parameters are drawn at random, and a world enters the benchmark only after audits confirm that the gold fix works, decoys are confounded rather than causal, the marker moves with the cause, and inert controls are inert.
The gold fix is found by the oracle's search over single controls and combinations of up to three, with dose refinement, and passes in every test world when regraded on fresh units.

Each domain has one shared pool of realistic names, and each world assigns roles to the names it draws at random, so no name predicts its role and domain knowledge is deliberately unreliable as a guide to causal structure.
In the world of \cref{fig:overview}, irrigation frequency runs at about 12.1 pulses per day, far from its best setpoint of 25.6.
The deviation, scaled by a per-unit susceptibility, drives a hidden micronutrient lockout through two hidden stages into leaf damage, and leaf necrosis is the measured marker.
Three treatments relieve the damage without removing the cause, three controls lower yield above a threshold dose, and canopy temperature and soil moisture correlate with yield only through a hidden planting cohort.
Setting irrigation frequency to 25.6 pulses per day raises yield for a typical unit from about 690 to about 1{,}040\,g/m$^2$/week.

The marker item passes if the named signal is a valid marker of the hidden cause, the decoy item is the Jaccard overlap with the world's confounded decoys, and each sign item checks one lever's effect on the true objective.
Signs are scored only over the world's levers, so an answer cannot pass by listing many controls as having no effect.

\section{Parser}
\label{app:parser}

In \query, over 31{,}252 query attempts, 93.8\% of requests by Claude agents and 79.4\% by other agents parse and validate; a rejected request uses one of the four turns but no data, and the agent can reformulate it.
In \repair, we re-resolved the 1{,}132 final answers of the first runs and of Claude Opus 5.5 on all test worlds with a different Claude model as the parser, and no fix verdict changed; re-resolving the causal accounts of a 72-answer sample changed no acceptance verdict.
A lexical parser without any LLM reproduces the original decision on 99.3\% of the 23{,}975 distinct experiment requests, because agents mostly name signals and controls as the situation report does.

\section{Models}
\label{app:models}

\Cref{tab:rep-models} lists the \repair models and their settings.
\query uses the same providers. Its agents sample at temperature 0.3 and the guess-only baseline at 0.1, except the Claude 5.5 and GPT-6 models on Bedrock, which reject an explicit temperature and run at the provider default.
The parser is Claude Opus 4.8 in \repair and Claude Opus 4.6 in \query.

\begin{table}[h]
\centering
\small
\caption{\repair models: identifier as called, maximum output tokens per reply, access route and sampling preset.}
\label{tab:rep-models}
\setlength{\tabcolsep}{3pt}
\newcommand{\repid}[1]{{\scriptsize\texttt{#1}}}
\begin{tabular}{@{}lllP{3.5cm}@{}}
\toprule
Model & Identifier & Max out & Access; sampling \\
\midrule
Claude Opus 5.5 & \repid{us.anthropic.claude-opus-5-5} & 128{,}000 & Bedrock; default \\
Claude Sonnet 5.5 & \repid{us.anthropic.claude-sonnet-5-5} & 128{,}000 & Bedrock; default \\
Claude Haiku 5.5 & \repid{us.anthropic.claude-haiku-5-5} & 32{,}768 & Bedrock; default \\
GPT-6.1 Sol & \repid{us.openai.gpt-6.1-sol} & 128{,}000 & Bedrock; default \\
GPT-6 Astra & \repid{us.openai.gpt-6-astra} & 128{,}000 & Bedrock; default \\
Gemini 3.1 Pro & \repid{gemini-3.1-pro-preview} & 32{,}768 & Google, OpenAI-compatible API; $T=1.0$, top-$p$ 0.95 \\
DeepSeek-V4-Pro & \repid{deepseek-v4-pro} & 32{,}768 & DeepSeek API; $T=1.0$, top-$p$ 0.95, thinking on, effort high \\
Qwen3.8-27B & \repid{Qwen/Qwen3.8-27B} & 32{,}768 & hosted OpenAI-compatible endpoint; $T=1.0$, top-$p$ 0.95, top-$k$ 20 \\
Qwen3.5-9B & \repid{Qwen3.5-9B} & 32{,}768 & local vLLM 0.19.1, thinking on; $T=1.0$, top-$p$ 0.95, top-$k$ 20 \\
Gemma 4 31B & \repid{gemma-4-31b-it} & 32{,}768 & local vLLM 0.27.1; $T=1.0$, top-$p$ 0.95, top-$k$ 64 \\
GPT-OSS-120B & \repid{openai.gpt-oss-120b-1:0} & 32{,}768 & Bedrock; default \\
Llama 4 Maverick & \repid{us.meta.llama4-maverick-17b-instruct-v1:0} & 8{,}192 & Bedrock (model maximum); default \\
\bottomrule
\end{tabular}
\end{table}

\newpage
\section{Protocol ablations}
\label{app:ablations}

Each ablation changes one thing and keeps the worlds, simulator, grader, parser and the rest of the prompt identical, and each is paired with the same model's first run on the evaluation set (\cref{tab:rep-ablations}).
Full history sends the whole conversation every turn, and the budget ablations set 30 or 8 experiments.
Repeated runs with the standard protocol show the run-to-run variation.

\begin{table}[h]
\centering
\footnotesize
\caption{Protocol ablations on the evaluation set, paired with each model's first run on the same worlds: \fixrate{} and acceptance (\%) in the first run and under the change, paired difference in points with a 95\% bootstrap interval over worlds, and exact McNemar $p$.}
\label{tab:rep-ablations}
\setlength{\tabcolsep}{3pt}
\begin{tabular}{@{}lcccccc@{}}
\toprule
 & \multicolumn{3}{c}{\fixrate{}} & \multicolumn{3}{c}{Acceptance} \\
\cmidrule(lr){2-4}\cmidrule(l){5-7}
Change, model & first $\to$ change & $\Delta$ [95\% CI] & $p$ & first $\to$ change & $\Delta$ [95\% CI] & $p$ \\
\midrule
\multicolumn{7}{@{}l}{\emph{Repeated run, standard protocol}} \\
Claude Opus 5.5, run 2 & 39 $\to$ 35 & $-4$ [$-15$, $+7$] & 0.75 & 17 $\to$ 17 & $+0$ [$-7$, $+7$] & 1 \\
Claude Opus 5.5, run 3 & 39 $\to$ 39 & $+0$ [$-9$, $+9$] & 1 & 17 $\to$ 19 & $+2$ [$-7$, $+11$] & 1 \\
GPT-6.1 Sol, run 2 & 50 $\to$ 35 & $-15$ [$-28$, $-4$] & 0.039 & 43 $\to$ 22 & $-20$ [$-33$, $-9$] & 0.003 \\
GPT-6.1 Sol, run 3 & 50 $\to$ 46 & $-4$ [$-15$, $+6$] & 0.73 & 43 $\to$ 31 & $-11$ [$-24$, $+2$] & 0.15 \\
GPT-6 Astra, run 2 & 50 $\to$ 59 & $+9$ [$-4$, $+22$] & 0.27 & 41 $\to$ 48 & $+7$ [$-4$, $+19$] & 0.34 \\
GPT-6 Astra, run 3 & 50 $\to$ 50 & $+0$ [$-11$, $+11$] & 1 & 41 $\to$ 39 & $-2$ [$-13$, $+9$] & 1 \\
\midrule
\multicolumn{7}{@{}l}{\emph{Full history}} \\
Claude Opus 5.5 & 39 $\to$ 52 & $+13$ [$+6$, $+22$] & 0.016 & 17 $\to$ 19 & $+2$ [$-6$, $+9$] & 1 \\
GPT-6.1 Sol & 50 $\to$ 52 & $+2$ [$-11$, $+15$] & 1 & 43 $\to$ 43 & $+0$ [$-13$, $+13$] & 1 \\
GPT-6 Astra & 50 $\to$ 65 & $+15$ [$+4$, $+26$] & 0.021 & 41 $\to$ 56 & $+15$ [$+4$, $+26$] & 0.021 \\
Claude Haiku 5.5 & 33 $\to$ 26 & $-7$ [$-20$, $+6$] & 0.39 & 11 $\to$ 13 & $+2$ [$-11$, $+13$] & 1 \\
Qwen3.8-27B & 15 $\to$ 33 & $+19$ [$+4$, $+33$] & 0.031 & 4 $\to$ 19 & $+15$ [$+6$, $+24$] & 0.008 \\
\midrule
\multicolumn{7}{@{}l}{\emph{Claude Opus 5.5: experiment budget}} \\
Budget 30 & 39 $\to$ 48 & $+9$ [$-2$, $+20$] & 0.18 & 17 $\to$ 30 & $+13$ [$+4$, $+24$] & 0.039 \\
Budget 8 & 39 $\to$ 26 & $-13$ [$-24$, $-2$] & 0.065 & 17 $\to$ 13 & $-4$ [$-13$, $+6$] & 0.69 \\
\bottomrule
\end{tabular}
\end{table}

\newpage
\section*{NeurIPS Paper Checklist}

\begin{enumerate}

\item {\bf Claims}
    \item[] Question: Do the main claims made in the abstract and introduction accurately reflect the paper's contributions and scope?
    \item[] Answer: \answerYes{}.
    \item[] Justification: The abstract and introduction state the benchmark design, its two suites, the evaluated models and agents, and the main findings; Section~\ref{sec:limitations} scopes the claims to synthetic hidden-model worlds and budgeted investigation.

\item {\bf Limitations}
    \item[] Question: Does the paper discuss the limitations of the work performed by the authors?
    \item[] Answer: \answerYes{}.
    \item[] Justification: Section~\ref{sec:limitations} discusses the scope (continuous SCMs and fully observed categorical PGMs, without temporal dynamics, measurement error or misspecification), the synthetic scenarios, what the milestone trace can and cannot attribute, and the risk of reading benchmark success as deployment readiness.

\item {\bf Theory assumptions and proofs}
    \item[] Question: For each theoretical result, does the paper provide the full set of assumptions and a complete (and correct) proof?
    \item[] Answer: \answerNA{}.
    \item[] Justification: The paper has no theoretical results. The causal models, intervention semantics, validation rules and oracle computations are part of the benchmark construction and are described in Section~\ref{sec:design} and Appendices~\ref{app:query} and~\ref{app:repair}.

    \item {\bf Experimental result reproducibility}
    \item[] Question: Does the paper fully disclose all the information needed to reproduce the main experimental results of the paper to the extent that it affects the main claims and/or conclusions of the paper (regardless of whether the code and data are provided or not)?
    \item[] Answer: \answerYes{}.
    \item[] Justification: Sections~\ref{sec:design} and~\ref{sec:protocol} specify the worlds, agent interface, budgets, parser, scoring and statistics, and the appendices give the remaining settings, including model identifiers and sampling settings (Appendix~\ref{app:models}). The released code runs the protocol of both suites.

\item {\bf Open access to data and code}
    \item[] Question: Does the paper provide open access to the data and code, with sufficient instructions to faithfully reproduce the main experimental results, as described in supplemental material?
    \item[] Answer: \answerYes{}.
    \item[] Justification: Code and data are available at \url{https://github.com/Rose-STL-Lab/ACED-Bench} under the MIT License: all benchmark worlds, the simulators, parsers, oracles and scorers, the agents, pinned dependencies, a Docker image, and a README with the commands that run the protocol of both suites. The world-generation scripts are not included.

\item {\bf Experimental setting/details}
    \item[] Question: Does the paper specify all the training and test details (e.g., data splits, hyperparameters, how they were chosen, type of optimizer) necessary to understand the results?
    \item[] Answer: \answerYes{}.
    \item[] Justification: Section~\ref{sec:protocol} and Appendices~\ref{app:models} and~\ref{app:ablations} give the evaluated models, agents, budgets, sampling settings, parser and scoring rules. There is no model training; the experiments evaluate LLM agents on fixed benchmark worlds.

\item {\bf Experiment statistical significance}
    \item[] Question: Does the paper report error bars suitably and correctly defined or other appropriate information about the statistical significance of the experiments?
    \item[] Answer: \answerYes{}.
    \item[] Justification: Section~\ref{sec:protocol} defines the statistics: 95\% bootstrap intervals over worlds for the main results (Figure~\ref{fig:leaderboard}, Table~\ref{tab:query_full}) and paired differences with exact McNemar tests for the protocol ablations (Table~\ref{tab:rep-ablations}).

\item {\bf Experiments compute resources}
    \item[] Question: For each experiment, does the paper provide sufficient information on the computer resources (type of compute workers, memory, time of execution) needed to reproduce the experiments?
    \item[] Answer: \answerYes{}.
    \item[] Justification: The experiments involve no training. Models are queried through hosted APIs, except two open-weight models on \repair that run on a local vLLM server (Appendix~\ref{app:models}); simulation, scoring and analysis run on a CPU.
    
\item {\bf Code of ethics}
    \item[] Question: Does the research conducted in the paper conform, in every respect, with the NeurIPS Code of Ethics \url{https://neurips.cc/public/EthicsGuidelines}?
    \item[] Answer: \answerYes{}.
    \item[] Justification: The work uses synthetic benchmark worlds and evaluates LLM agents; it collects no personal data. The only human input is a blind annotation of 48 anonymised evidence-ledger turns for validation (Appendix~\ref{app:query_ledger}).

\item {\bf Broader impacts}
    \item[] Question: Does the paper discuss both potential positive societal impacts and negative societal impacts of the work performed?
    \item[] Answer: \answerYes{}.
    \item[] Justification: Section~\ref{sec:limitations} frames the positive impact as sharper diagnostic evaluation of AI co-scientist systems and the negative risk as overinterpreting synthetic-benchmark scores as deployment readiness, especially where real-world data quality, fairness, privacy, and intervention semantics are contested.
    
\item {\bf Safeguards}
    \item[] Question: Does the paper describe safeguards that have been put in place for responsible release of data or models that have a high risk for misuse (e.g., pre-trained language models, image generators, or scraped datasets)?
    \item[] Answer: \answerNA{}.
    \item[] Justification: The paper does not release pretrained models, image generators, scraped datasets, or other high-misuse-risk assets. The released assets are synthetic benchmark worlds and evaluation code.

\item {\bf Licenses for existing assets}
    \item[] Question: Are the creators or original owners of assets (e.g., code, data, models), used in the paper, properly credited and are the license and terms of use explicitly mentioned and properly respected?
    \item[] Answer: \answerYes{}.
    \item[] Justification: The existing assets are hosted LLM services, used under their providers' terms (Appendix~\ref{app:models}), and open-source Python packages listed in the release's pinned requirements. The release does not redistribute external datasets, model weights or third-party media.

\item {\bf New assets}
    \item[] Question: Are new assets introduced in the paper well documented and is the documentation provided alongside the assets?
    \item[] Answer: \answerYes{}.
    \item[] Justification: The benchmark worlds and code are documented in the paper and the repository README and released under the MIT License. The release contains the worlds, simulators, oracles, scorers, agents and tests; the world-generation scripts are not included.

\item {\bf Crowdsourcing and research with human subjects}
    \item[] Question: For crowdsourcing experiments and research with human subjects, does the paper include the full text of instructions given to participants and screenshots, if applicable, as well as details about compensation (if any)? 
    \item[] Answer: \answerNA{}.
    \item[] Justification: The paper involves no crowdsourcing or human-subject experiment. The only human input is a blind annotation of 48 anonymised agent turns used to validate the evidence-ledger labels (Appendix~\ref{app:query_ledger}); it involves no personal data.

\item {\bf Institutional review board (IRB) approvals or equivalent for research with human subjects}
    \item[] Question: Does the paper describe potential risks incurred by study participants, whether such risks were disclosed to the subjects, and whether Institutional Review Board (IRB) approvals (or an equivalent approval/review based on the requirements of your country or institution) were obtained?
    \item[] Answer: \answerNA{}.
    \item[] Justification: No human-subject research was conducted (see the previous item), so IRB review does not apply.

\item {\bf Declaration of LLM usage}
    \item[] Question: Does the paper describe the usage of LLMs if it is an important, original, or non-standard component of the core methods in this research? Note that if the LLM is used only for writing, editing, or formatting purposes and does \emph{not} impact the core methodology, scientific rigor, or originality of the research, declaration is not required.
    \item[] Answer: \answerYes{}.
    \item[] Justification: LLMs are the evaluated agents, draft the natural-language surface of the \query worlds, serve as the fixed parser and answer extractor, and annotate the evidence ledger (Sections~\ref{sec:design} and~\ref{sec:protocol}, Appendix~\ref{app:query_ledger}).

\end{enumerate}

\end{document}